\documentclass{article}
\usepackage{color,soul}
\usepackage{microtype}
\usepackage{graphicx}
\usepackage{caption}
\usepackage{subcaption}
\usepackage{booktabs} 
\usepackage{amsmath}
\usepackage{bbm}
\usepackage{multirow}
\usepackage{algorithm2e}
\usepackage{glossaries}
\makeatletter
\let\oldmakefirstuc\makefirstuc
\renewcommand*{\makefirstuc}[1]{%
  \def\gls@add@space{}%
  \mfu@capitalisewords#1 \@nil\mfu@endcap
}
\def\mfu@capitalisewords#1 #2\mfu@endcap{%
  \def\mfu@cap@first{#1}%
  \def\mfu@cap@second{#2}%
  \gls@add@space
  \oldmakefirstuc{#1}%
  \def\gls@add@space{ }%
  \ifx\mfu@cap@second\@nnil
    \let\next@mfu@cap\mfu@noop
  \else
    \let\next@mfu@cap\mfu@capitalisewords
  \fi
  \next@mfu@cap#2\mfu@endcap
}
\makeatother

\usepackage{hyperref}
\usepackage[capitalise]{cleveref}

\usepackage[accepted]{icml2022}

\newacronym{mhrot}{MHRot}{multi-head RotNet}
\newacronym{ntl}{NTL}{anomaly detection using neural transformations}
\newacronym{icl}{ICL}{anomaly detection with internal contrastive learning}
\newacronym{loe}{LOE}{latent outlier exposure}

\newcommand{\Ln}{{\mathcal{L}^\theta_n}}
\newcommand{\La}{{\mathcal{L}^\theta_a}}
\newcommand{\x}{{\bf x}}
\newcommand{\y}{{\bf y}}

\icmltitlerunning{Latent Outlier Exposure for Anomaly Detection with Contaminated Training Data}

\begin{document}

\twocolumn[
\icmltitle{Latent Outlier Exposure for Anomaly Detection with Contaminated Data}



\icmlsetsymbol{equal}{*}

\begin{icmlauthorlist}
\icmlauthor{Chen Qiu}{equal,bcai,tuk}
\icmlauthor{Aodong Li}{equal,uci}
\icmlauthor{Marius Kloft}{tuk}
\icmlauthor{Maja Rudolph}{bcai}
\icmlauthor{Stephan Mandt}{uci}
\end{icmlauthorlist}

\icmlaffiliation{bcai}{Bosch Center for Artificial Intelligence}
\icmlaffiliation{tuk}{TU Kaiserslautern, Germany}
\icmlaffiliation{uci}{UC Irvine, USA}

\icmlcorrespondingauthor{Chen Qiu}{chen.qiu@de.bosch.com}
\icmlcorrespondingauthor{Stephan Mandt}{mandt@uci.edu}

\icmlkeywords{Machine Learning, ICML}

\vskip 0.3in
]



\printAffiliationsAndNotice{\icmlEqualContribution} 

\begin{abstract}
Anomaly detection aims at identifying data points that show systematic deviations from the majority of data in an unlabeled dataset. A common assumption is that clean training data (free of anomalies) is available, which is often violated in practice. We propose a  strategy for training an anomaly detector in the presence of unlabeled anomalies that is compatible with a broad class of models. The idea is to jointly infer binary labels to each datum (normal vs. anomalous) while updating the model parameters. Inspired by outlier exposure \citep{hendrycks2018deep} that considers synthetically created, labeled anomalies, we thereby use a combination of two losses that share parameters: one for the normal and one for the anomalous data. We then iteratively proceed with block coordinate updates on the parameters and the most likely (latent) labels. 
Our experiments with several backbone models on three image datasets, 30 tabular data sets, and a video anomaly detection benchmark showed consistent and significant improvements over the baselines.
\end{abstract}

\section{Introduction}
From industrial fault detection to medical image analysis or financial fraud prevention: Anomaly detection---the task of automatically identifying anomalous data instances without being explicitly taught how anomalies may look like---is critical in industrial and technological applications. 
    
The common approach in deep anomaly detection is to first train a neural network on a large dataset of ``normal" samples minimizing some loss function (such as a deep one-class classifier~\citep{ruff2018deep}) and to then construct an anomaly score from the output of the neural network (typically based on the training loss).  Anomalies are then identified as data points with larger-than-usual anomaly scores and obtained by thresholding the score at particular~values. 
   
A standard assumption in this approach is that clean training data are available to teach the model what ``normal" samples look like \citep{ruff2021unifying}. In reality, this assumption is often violated: datasets are frequently large and uncurated and may already contain some of the anomalies one is hoping to find. For example, a dataset of medical images may already contain cancer images, or datasets of financial transactions could already contain unnoticed fraudulent activity. Naively training an unsupervised anomaly detector on such data may suffer from degraded performance. 
   
In this paper, we introduce a new unsupervised approach to training anomaly detectors on a corrupted dataset. Our approach uses a combination of two coupled losses to extract learning signals from both normal and anomalous data. We stress that these losses do not necessarily have a probabilistic interpretation; rather, many recently proposed self-supervised auxiliary losses can be used \citep{ruff2018deep,hendrycks2019using,qiu2021neural,shenkar2022anomaly}. In order to decide which of the two loss functions to activate for a given datum (normal vs. abnormal), we use a binary latent variable that we jointly infer while updating the model parameters. Training the model thus results in a joint optimization problem over continuous model parameters and binary variables that we solve using alternating updates. During testing, we can use threshold only one of the two loss functions to identify anomalies in constant time.

Our approach can be applied to a variety of anomaly detection loss functions 
and data types, as we demonstrate on tabular, image, and video data. Beyond detection of entire anomalous images, we also consider the problem of anomaly segmentation  which is concerned with finding anomalous regions within an image. Compared to established baselines that either ignore the anomalies or try to iteratively remove them~\citep{yoon2021self}, our approach yields significant performance improvements in all cases. 

The paper is structured as follows. In \Cref{sec:related}, we discuss related work. In \Cref{sec:method}, we introduce our main algorithm, including the involved losses and optimization procedure. Finally, in \Cref{sec:experiments}, we discuss experiments on both image and tabular data and discuss our findings in \Cref{sec:conclusion} \footnote{Code is available at \url{https://github.com/boschresearch/LatentOE-AD.git}}. 
 

\section{Related Work}
\label{sec:related}
We divide our related work into methods for deep anomaly detection, learning on incomplete or contaminated data, and training anomaly detectors on contaminated data. 
\paragraph{Deep anomaly detection.}
Deep learning has played an important role in recent advances in anomaly detection. For example, \citet{ruff2018deep} have improved the anomaly detection accuracy of one-class classification \citep{scholkopf2001estimating} by combining it with a deep feature extractor, both in the unsupervised and the semi-supervised setting \citep{ruff2019deep}. An alternative strategy to combine deep learning with one-class approaches is to train a one-class SVM on pretrained self-supervised features \citep{sohn2020learning}. Indeed, self-supervised learning has influenced deep anomaly detection in a number of ways: The self-supervised criterion for training a deep feature extractor can be used directly to score anomalies \citep{golan2018deep,bergman2020classification}. Using a \gls{mhrot}, \citet{hendrycks2019using} improve self-supervised anomaly detection by solving multiple classification tasks. 
For general data types beyond images, \gls{ntl}~\citep{qiu2021neural,qiu2022raising} learns the transformations for the self-supervision task and achieves solid detection accuracy. \citet{schneider2022detecting} combine \gls{ntl} with representation learning for detecting anomalies within time series.
On tabular data, \gls{icl}~\citep{shenkar2022anomaly} learns feature relations as a self-supervised learning task.
Other classes of deep anomaly detection includes autoencoder variants \citep{principi2017acoustic,zhou2017anomaly,chen2018unsupervised} and density-based models \citep{schlegl2017unsupervised, deecke2018image}.

All these approaches assume a training dataset of ``normal'' data. 
However, in many practical scenarios there will be unlabeled anomalies hidden in the training data. \citet{wang2019effective,huyan2021unsupervised} have shown that anomaly detection accuracy deteriorates when the training set is contaminated. Our work provides a training strategy to deal with contamination. 

\paragraph{Anomaly Detection on contaminated training data.} 
A common strategy to deal with contaminated training data is to hope that the contamination ratio is low 
and that the anomaly detection method will exercise {\em inlier priority} \citep{wang2019effective}. Throughout our paper, we refer to the strategy of blindly training an anomaly detector as if the training data was clean as ``\textit{Blind}'' training. 
\citet{yoon2021self} have proposed a data refinement strategy that removes potential anomalies from the training data. Their approach, which we refer to as ``{\em Refine}'', employs an ensemble of one-class classifiers to iteratively weed out anomalies and then to continue training on the refined dataset. Similar data refinement strategy are also combined with latent SVDD \citep{gornitz2014learning} or autoencoders for anomaly detection \citep{xia2015learning,beggel2019robust}.
However, these methods fail to exploit the insight of outlier exposure
\citep{hendrycks2018deep} that anomalies provide a valuable training signal.
\citet{zhou2017anomaly} used a robust autoencoder for identifying anomalous training data points, but their approach requires training a new model for identifying anomalies, which is impractical in most setups.  
\citet{hendrycks2018deep} 
propose to artificially contaminate the training data with samples from a related domain which can then be considered anomalies. While outlier exposure assumes labeled anomalies, our work aims at exploiting unlabeled anomalies in the training data.
Notably, \citet{pang2020self} have used an iterative scheme to detect abnormal frames in video clips, and \citet{feng2021mist} extend it to supervised video anomaly detection.
Our work is more general and provides a principled way to improve the training strategy of all approaches mentioned in the paragraph ``deep anomaly detection'' when the training data is likely contaminated.

\section{Method}
\label{sec:method}
We will start by describing the mathematical foundations of our method. We will then describe our learning algorithm as a block coordinate descent algorithm, providing a theoretical convergence guarantee. Finally, we describe how our approach is applicable in the context of various state-of-the-art deep anomaly detection methods. 

\subsection{Problem Formulation}

\paragraph{Setup.} In this paper, we study the problem of unsupervised (or self-supervised) anomaly detection. We consider a data set of samples $\x_i$; these could either come from a data distribution of ``normal" samples, or could  otherwise come from an unknown corruption process and thus be considered as ``anomalies". For each datum $\x_i$, let $y_i=0$ if the datum is normal, and $y_i=1$ if it is anomalous. We assume that these binary labels are unobserved, both in our training and test sets, and have to be inferred from the data.

In contrast to most anomaly detection setups, we assume that our dataset is \emph{corrupted by anomalies}. That means, we assume that a fraction $(1-\alpha)$ of the data is normal, while its complementary fraction $\alpha$ is anomalous. This corresponds to a more challenging (but arguably more realistic) anomaly detection setup since the training data cannot be assumed to be normal. We treat the assumed contamination ratio $\alpha$ as a hyperparameter in our approach and denote $\alpha_0$ as the ground truth contamination ratio where needed. Note that an assumed contamination ratio is a common hyperparameter in many robust algorithms \citep[e.g.,][]{huber1992robust,huber2011robust}, and we test the robustness of our approach w.r.t. this parameter in \Cref{sec:experiments}.

Our goal is to train a (deep) anomaly detection classifier on such corrupted data based on self-supervised or unsupervised training paradigms (see related work). The challenge thereby is to simultaneously infer the binary labels $y_i$ during training while optimally exploiting this information for training an anomaly detection model. 

\paragraph{Proposed Approach.}
We consider two losses. Similar to most work on deep anomaly detection, we consider a loss function $\Ln(\x)\equiv {\cal L}_n(f_\theta(\x))$ that we aim to minimize over ``normal" data. The function $f_\theta(\x)$ is used to extract features from $\x$, typically based on a self-supervised auxiliary task, see \Cref{sec:examples} for examples. When being trained on only normal data, the trained loss will yield lower values for normal than for anomalous data so that it can be used to construct an anomaly score. 

In addition, we also consider a second loss for anomalies $\La(\x)\equiv {\cal L}_a (f_\theta(\x))$ (the feature extractor $f_\theta(\x)$ is shared). Minimizing this loss on only anomalous data will result in low loss values for anomalies and larger values for normal data. The anomaly loss is designed to have opposite effects as the loss function $\Ln(\x)$. For example, if $\Ln(\x)=||f_\theta(\x)-{\bf c}||^2$ as in Deep SVDD \citep{ruff2018deep} (thus pulling normal data points towards their center), we define $\La(\x) = 1/||f_\theta(\x)-{\bf c}||^2$ (pushing abnormal data away from it) as in \citep{ruff2019deep}. 

Temporarily assuming that all assignment variables $\y$ were known, consider the joint loss function,
\begin{align}
\label{eq:loe-loss}
    {\cal L}(\theta, \y) = \sum_{i=1}^N (1-y_i) \Ln(\x_i) + y_i \La(\x_i).
\end{align}
This equation resembles the log-likelihood of a probabilistic mixture model, but note that $\Ln(\x_i)$ and $\La(\x_i)$ are not necessarily data log-likelihoods; rather, self-supervised auxiliary losses can be used and often perform better in practice \citep{ruff2018deep, qiu2021neural, nalisnick2018deep}. 

Optimizing Eq.~\ref{eq:loe-loss} over its parameters $\theta$ yields a better anomaly detector than $\Ln$ trained in isolation. By construction of the anomaly loss $\La$, the known anomalies provide an additional training signal to $\Ln$: due to parameter sharing, the labeled anomalies teach $\Ln$ where \emph{not} to expect normal data in feature space. This is the basic idea of Outlier Exposure \citep{hendrycks2018deep}, which constructs artificial \emph{labeled} anomalies for enhanced detection performance. 

Different from Outlier Exposure, we assume that the set of $y_i$ is unobserved, hence \emph{latent}. We therefore term our approach of jointly inferring latent assignment variables $\y$ and learning parameters $\theta$ as \emph{\Gls{loe}}. We show that it leads to competitive performance on training data corrupted by outliers.

\subsection{Optimization problem}
\label{sec:optimization}

\paragraph{``Hard" Latent Outlier Exposure (\gls{loe}$_H$).}
In \gls{loe}, we seek to both optimize both losses' shared parameters $\theta$ while also optimizing the most likely assignment variables $y_i$. Due to our assumption of having a fixed rate of anomalies $\alpha$ in the training data, we introduce a constrained set:
\begin{align}
\label{eq:label-constraint}
    {\cal Y} = \{\y \in \{0,1\}^N: \sum_{i=1}^N y_i = \alpha N\}.
\end{align}
The set describes a ``hard" label assignment; hence the name ``Hard \gls{loe}", which is the default version of our approach.  
\Cref{sec:extension} describes an extension with ``soft" label assignments. Note that we require $\alpha N$ to be an integer. 

Since our goal is to use the losses $\Ln$ and $\La$ to identify and score anomalies, we seek $\Ln(\x_i) - \La(\x_i)$ to be large for anomalies, and $\La(\x_i) - \Ln(\x_i)$ to be large for normal data. Assuming these losses to be optimized over $\theta$, our best guess to identify anomalies is to minimize \cref{eq:loe-loss} over the assignment variables $\y$. Combining this with the constraint  (\cref{eq:label-constraint}) yields the following minimization problem:
\begin{align}
    \label{eq:optimization-problem}
    \min_\theta \min_{\y \in {\cal Y}} \;  {\cal L}(\theta,\y).
\end{align}
As follows, we describe an efficient optimization procedure for the constraint optimization problem. 

\paragraph{Block coordinate descent.} 
The constraint discrete optimization problem has an elegant solution. 

To this end, we consider a sequence of parameters $\theta^t$ and labels $\y^t$ and proceed with alternating updates. To update $\theta$, we simply fix $\y^t$ and minimize ${\cal L}(\theta, \y^t)$ over $\theta$. In practice, we perform a single gradient step (or stochastic gradient step, see below), yielding a partial update. 

To update $\y$ given $\theta^t$, we minimize the same function subject to the constraint (\cref{eq:label-constraint}). To this end, we define training anomaly scores,
\begin{align}
    S_i^{train} = \Ln(\x_i) - \La(\x_i). 
\end{align}
These scores quantify the effect of $y_i$ on minimizing \cref{eq:loe-loss}. We rank these scores and assign the $(1-\alpha)$-quantile of the associated labels $y_i$ to the value $0$, and the remainder to the value $1$. This minimizes the loss function subject to the label constraint. We discuss the sensitivity of our approach to the assumed rate of anomalies $\alpha$ in our experiments section. We stress that our testing anomaly scores will be different (see \Cref{sec:extension}). 


Assuming that all involved losses are bounded from below, the block coordinate descent converges to a local optimum since every update improves the loss. 

\paragraph{Stochastic optimization.} In practice, we perform stochastic gradient descent on \cref{eq:loe-loss} based on mini-batches. For simplicity and memory efficiency, we impose the label constraint \cref{eq:label-constraint} on each mini-batch and optimize $\theta$ and $\y$ in the same alternating fashion. The induced bias vanishes for large mini-batches. In practice, we found that this approach leads to satisfying results\footnote{Note that an exact mini-batch version of the optimization problem in \cref{eq:optimization-problem} would also be possible, requiring memorization of $\y$ for the whole data set.}.

\Cref{alg:loe} summarizes our approach. 
\RestyleAlgo{ruled}
\SetKwInput{kwData}{Data}
\SetKwInput{kwModel}{Model}
\SetKwInput{kwInput}{Input}
\begin{algorithm}[t]
\caption{Training process of \gls{loe}}
\label{alg:loe}
\kwInput{Contaminated training set $\mathcal{D}$ ($\alpha_0$ anomaly rate)\\
$\qquad \quad$hyperparamter $\alpha$}
\kwModel{Deep anomaly detector with parameters $\theta$}
\ForEach{Epoch}{
\ForEach{Mini-batch $\mathcal{M}$}{
Calculate the anomaly score $S_i^{train}$ for $\x_i \in \mathcal{M}$\\
Estimate the label $y_i$ given $S_i^{train}$ and $\alpha$ \\ 
Update the parameters $\theta$ by minimizing ${\cal L}(\theta, \y)$

}
}
\end{algorithm}
\vspace{-5pt}

\subsection{Model extension and anomaly detection}
\label{sec:extension}
We first discuss an important extension of our approach and then discuss its usage in anomaly detection.

\paragraph{``Soft'' Latent Outlier Exposure (\gls{loe}$_S$).}
In practice, the block coordinate descent procedure can be overconfident in assigning $\y$, leading to suboptimal training. To overcome this problem, we also propose a \emph{soft} anomaly scoring approach that we term \emph{Soft} \gls{loe}.  Soft \gls{loe} is very simply implemented by a modified constraint set:
\begin{align}
    {\cal Y}' = \{\y \in \{0,0.5\}^N: \sum_{i=1}^N y_i = 0.5\alpha N\}.
\end{align}
Everything else about the model's training and testing scheme remains the same. 

The consequence of an identified anomaly $y_i=0.5$ is that we minimize an equal combination of both losses, $0.5(\Ln(\x_i) + \La(x_i))$. The interpretation is that the algorithm is uncertain about whether to treat $\x_i$ as a normal or anomalous data point and treats both cases as equally likely. A similar weighting scheme has been proposed for supervised learning in the presence of unlabeled examples \citep{lee2003learning}. In practice, we found the soft scheme to sometimes outperform the hard one (see \Cref{sec:experiments}). 

\paragraph{Anomaly Detection.}
In order to use our approach for finding anomalies in a test set, we could in principle proceed as we did during training and infer the most likely labels as described in \Cref{sec:optimization}. However, in practice we may not want to assume to encounter the same kinds of anomalies that we encountered during training. Hence, we refrain from using $\La$ during testing and score anomalies using only $\Ln$. Note that due to parameter sharing, training $\La$ jointly with $\Ln$ has already led to the desired information transfer between both losses.

Testing is the same for both ``soft" \gls{loe} (\Cref{sec:optimization}) and ``hard" \gls{loe} (\Cref{sec:extension}). We define our testing anomaly score in terms of the ``normal" loss function,
\begin{align}
    S^{test}_i = \Ln(\x_i).
\end{align}


\subsection{Example loss functions}
\label{sec:examples}
As follows, we review several loss functions that are compatible with our approach. We consider three advanced classes of self-supervised anomaly detection methods. 
These methods are i) \gls{mhrot} \citep{hendrycks2019using}, ii) \gls{ntl} \citep{qiu2021neural}, and iii) \gls{icl} \citep{shenkar2022anomaly}. While no longer being considered as a competitive baseline, we also consider deep SVDD for visualization due to its simplicity.

\paragraph{Multi-Head RotNet (MHRot).}
\gls{mhrot} \citep{hendrycks2019using} learns a multi-head classifier $f_\theta$ to predict the applied image transformations including rotation, horizontal shift, and vertical shift. We denote $K$ combined transformations as $\{T_{1},...,T_{K}\}$.
The classifier has three softmax heads, each for a classification task $l$, modeling the prediction distribution of a transformed image $p^l(\cdot | f_\theta, T_k(\x))$ (or $p_k^l(\cdot|\x)$ for brevity). 
Aiming to predict the correct transformations for normal samples, we maximize the log-likelihoods of the ground truth label $t_k^l$ for each transformation and each head;
for anomalies, we make the predictions evenly distributed by minimizing the cross-entropy from a uniform distribution $\mathcal{U}$ to the prediction distribution, resulting in 
\begin{align*}
     \Ln(\x)&:=-\textstyle\sum_{k=1}^{K}\sum_{l=1}^3  \log p_k^l(t_k^l|\x),\nonumber\\ \La(\x)&:=\textstyle\sum_{k=1}^{K}\sum_{l=1}^3 \text{CE}(\mathcal{U}, p_k^l(\cdot|\x))\nonumber
\end{align*}
\paragraph{Neural Transformation Learning (NTL).}
Rather than using hand-crafted transformations, \gls{ntl} learns $K$ neural transformations $\{T_{\theta,1},...,T_{\theta,K}\}$ and an encoder $f_\theta$ parameterized by $\theta$ from data and uses the learned transformations to detect anomalies. Each neural transformation generates a view $\x_k = T_{\theta,k}(\x)$ of sample $\x$. 
For normal samples, \gls{ntl} encourages each transformation to be similar to the original sample and to be dissimilar from other transformations. To achieve this objective, \gls{ntl} maximizes the normalized probability $p_k=h(\x_k, \x)\big/\big(h(\x_k, \x)+\sum_{l\neq k}h(\x_k, \x_l)\big)$ for each view where $h(\mathbf{a},\mathbf{b}) = \exp (\mathrm{cos}(f_\theta(\mathbf{a}), f_\theta(\mathbf{b}))/ \tau)$ measures the similarity of two views \footnote{
  where $\tau$ is the temperature and $\mathrm{cos}(\mathbf{a},\mathbf{b}) := \mathbf{a}^T \mathbf{b}/ \|\mathbf{a}\| \|\mathbf{b}\|$
}. For anomalies, we ``flip” the objective for normal samples: the model instead pulls the transformations close to each other and pushes them away from the original view, resulting in
\begin{align*}
    \Ln(\x):= -\sum_{k=1}^K \log p_k, \quad \La(\x):= -\sum_{k=1}^K \log (1-p_k).
\end{align*}

\paragraph{Internal Contrastive Learning (ICL).}
\gls{icl} is a state-of-the-art \textit{tabular} anomaly detection method \citep{shenkar2022anomaly}. Assuming that the relations between a subset of the features (table columns) and the complementary subset are class-dependent, \gls{icl} is able to learn an anomaly detector by discovering the feature relations for a specific class. 
With this in mind, \gls{icl} learns to maximize the mutual information between the two complementary feature subsets, $a(\x)$ and $b(\x)$, in the embedding space. The maximization of the mutual information is equivalent to minimizing a contrastive loss $\Ln(\x):= -\sum_{k=1}^K \log p_k$ on normal samples with $p_k=h(a_k(\x), b_k(\x))\big/\sum_{l=1}^K h(a_l(\x), b_k(\x))$ where $h(a,b) = \exp (\mathrm{cos}(f_\theta(a), g_\theta(b))/ \tau)$ measures the similarity of two feature subsets in the embedding space of two encoders $f_\theta$ and $g_\theta$. For anomalies, we flip the objective as $\La(\x):= -\sum_{k=1}^K \log (1-p_k)$.
\section{Experiments}
\label{sec:experiments}
We evaluate our proposed methods and baselines for unsupervised anomaly detection tasks on different data types: synthetic data, tabular data, images, and videos.
The data are contaminated with different anomaly ratios.
Depending on the data, we study our method in combination with specific backbone models. \Gls{mhrot} applies only to images and \gls{icl} to tabular data. \gls{ntl} can be applied to all data types. 

We have conducted extensive experiments on image, tabular, and video data. For instance, we evaluate our methods on all $30$ tabular datasets of \citet{shenkar2022anomaly}. Our proposed method sets a new state-of-the-art on most datasets. In particular, we show that our method gives robust results even when the contamination ratio is unknown. 



\subsection{Toy Example}
\begin{figure*}[t!]
    \centering
    	\begin{subfigure}[b]{0.33\linewidth}
		\includegraphics[width=\linewidth]{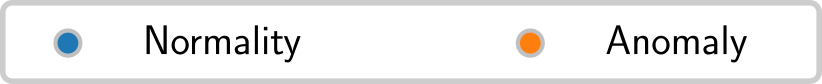}
	\end{subfigure}	\\
	\begin{subfigure}[b]{0.19\linewidth}
		\includegraphics[width=\linewidth]{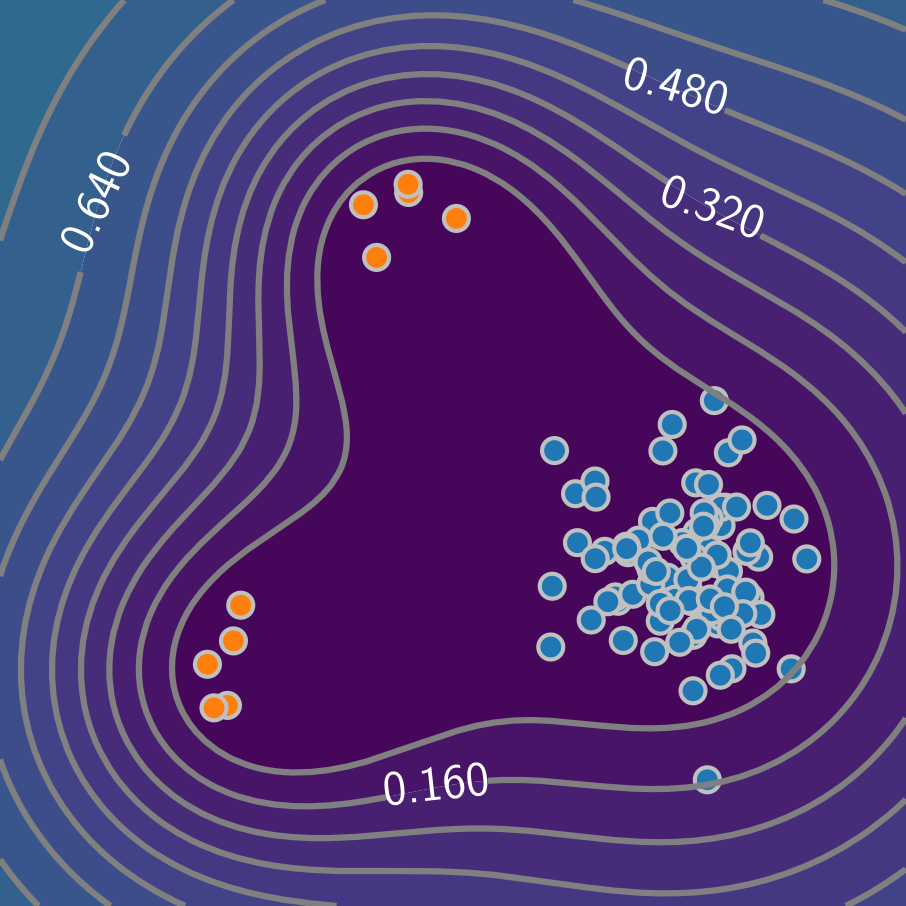}
	\caption{Blind}
	\label{fig:toy_blind}
	\end{subfigure}
		\begin{subfigure}[b]{0.19\linewidth}
		\includegraphics[width=\linewidth]{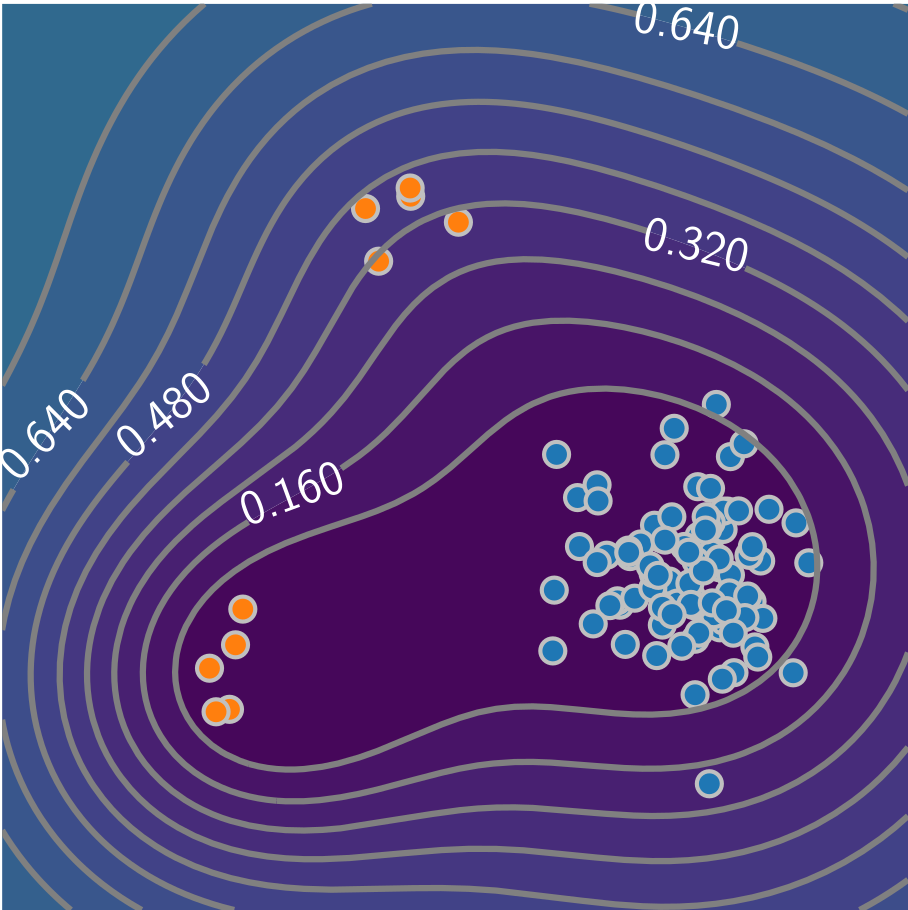}
	\caption{Refine}
	\label{fig:toy_refine}
	\end{subfigure}
	\begin{subfigure}[b]{0.19\linewidth}
		\includegraphics[width=\linewidth]{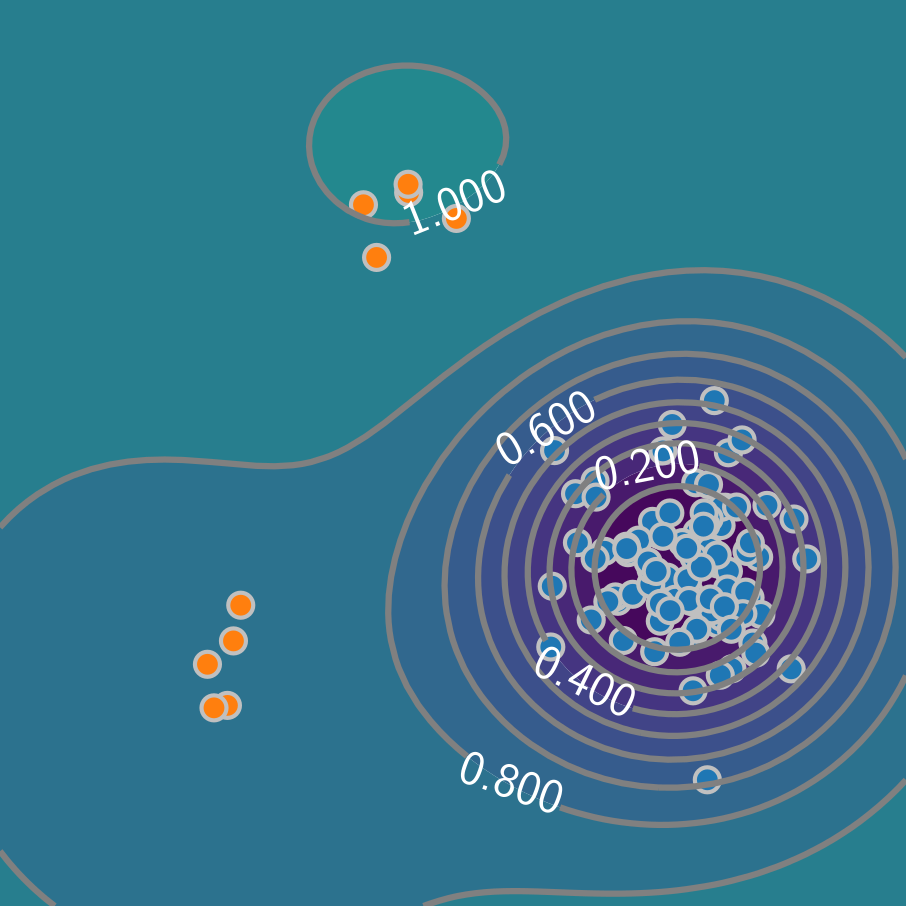}
    \caption{LOE$_{S}$}
    \label{fig:toy_soft}
	\end{subfigure}
	\begin{subfigure}[b]{0.19\linewidth}
		\includegraphics[width=\linewidth]{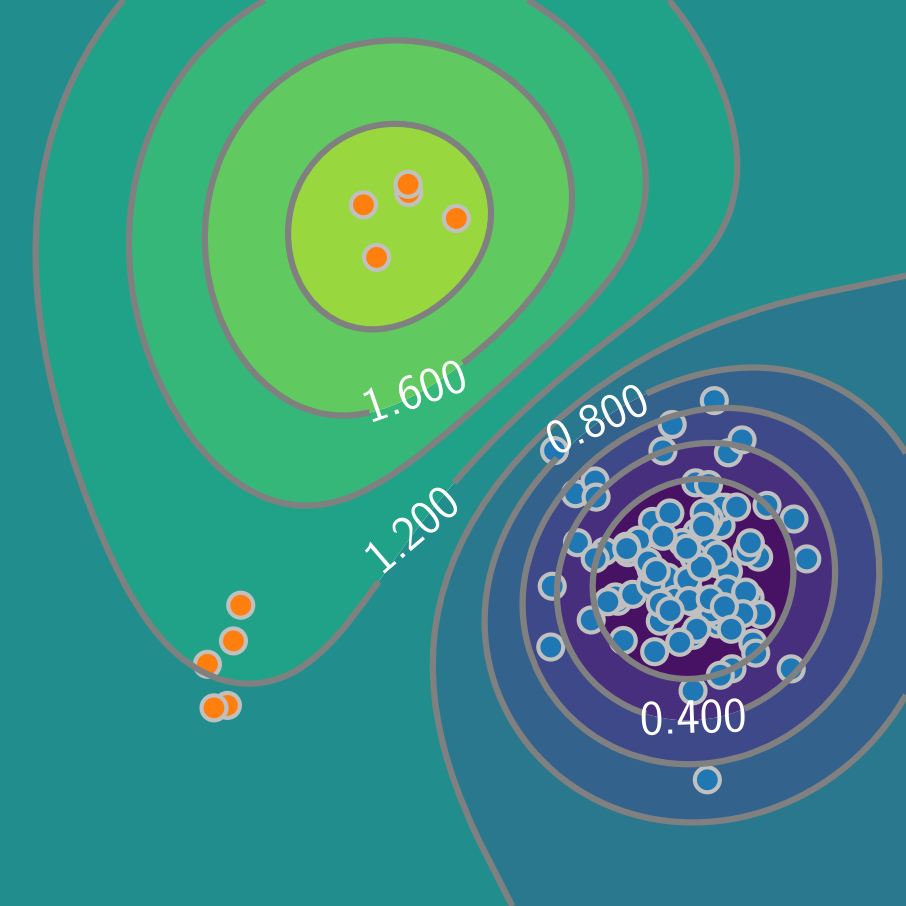}
    \caption{LOE$_{H}$}
    \label{fig:toy_hard}
	\end{subfigure}
		\begin{subfigure}[b]{0.19\linewidth}
		\includegraphics[width=\linewidth]{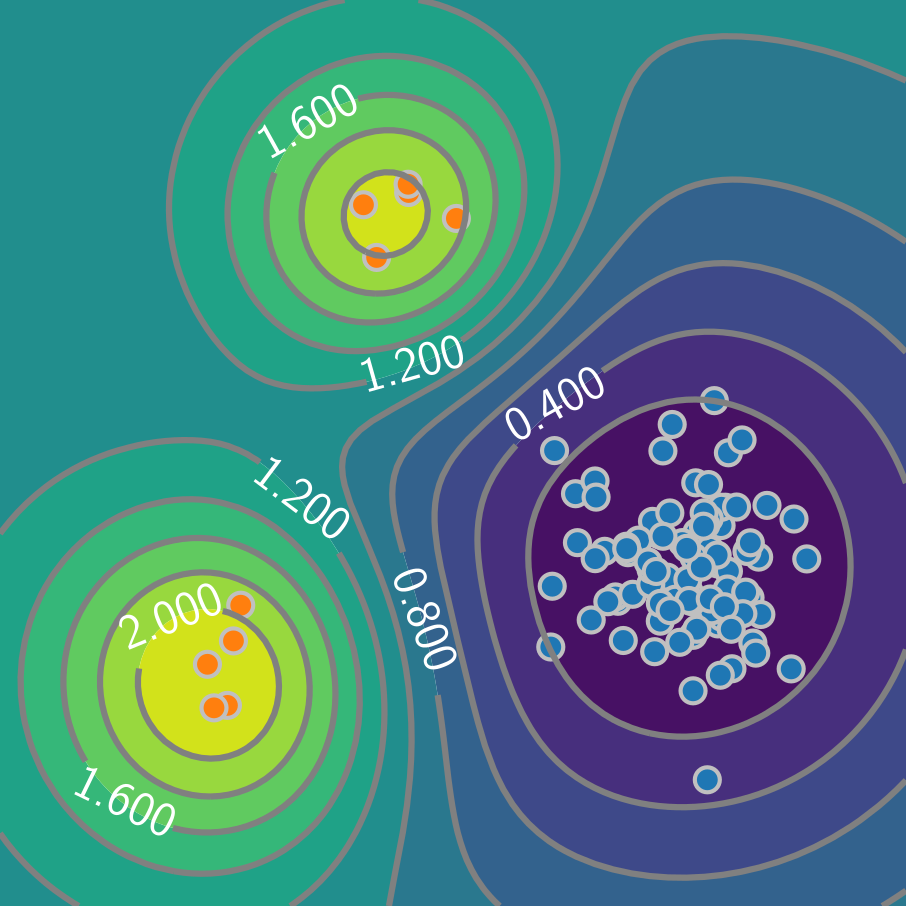}
    \caption{G-truth}
    \label{fig:toy_truth}
	\end{subfigure}
		\begin{subfigure}[b]{0.0255\linewidth}
		\includegraphics[width=\linewidth]{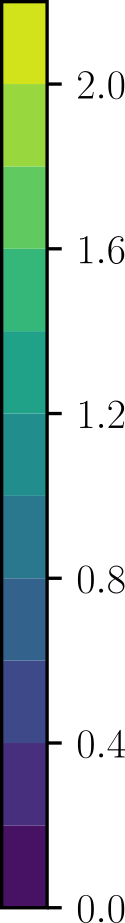}
	\caption*{}
	\end{subfigure}
	\\
    \caption{Deep SVDD trained on 2D synthetic contaminated data (see main text) trained with different methods: \textbf{(a)} ``Blind'' (treats all data as normal), \textbf{(b)} ``Refine'' (filters out some anomalies), \textbf{(c)} LOE$_S$ (proposed, assigns soft labels to anomalies),  \textbf{(d)} LOE$_H$ (proposed, assigns hard labels), \textbf{(e)} supervised anomaly detection with ground truth labels (for reference). \gls{loe} leads to improved region boundaries.} 
    \label{fig:toy_data}
\end{figure*}
We first analyze the methods in a controlled setup on a synthetic data set. 
For the sake of visualization, we created a 2D contaminated data set with a three-component Gaussian mixture. One larger component is used to generate normal samples, while the two smaller components are used to generate the anomalies contaminating the data (see \cref{fig:toy_data}). For simplicity, the backbone anomaly detector is the deep one-class classifier~\citep{ruff2018deep} with radial basis functions. Setting the contamination ratio to $\alpha_0=\alpha=0.1$, we compare the baselines ``Blind'' and ``Refine'' (described in \Cref{sec:related}, detailed in \Cref{app:baseline}) with the proposed \gls{loe}$_H$ and \gls{loe}$_S$ (described in \Cref{sec:method}) and the theoretically optimal \textit{G-truth} method (which uses the ground truth labels). We defer all further training details to \Cref{app:toy}.

\cref{fig:toy_data} shows the results (anomaly-score contour lines after training). 
With more latent anomaly information exploited from (a) to (e), the contour lines become increasingly accurate. While (a) ``Blind'' erroneously treats all anomalies as normal, (b) ``Refine'' improves by filtering out some anomalies. (c) \gls{loe}$_S$ and (d) \gls{loe}$_H$ use the anomalies, resulting in a clear separation of anomalies and normal data. \gls{loe}$_H$ leads to more pronounced boundaries than \gls{loe}$_S$, but it is at risk of overfitting, especially when normal samples are incorrectly detected as anomalies  (see our experiments below).  A supervised model with ground-truth labels (``G-truth'') approximately recovers the true contours.

\subsection{Experiments on Image Data}
Anomaly detection on images is especially far developed. We demonstrate \gls{loe}'s benefits when applied to two leading image anomaly detectors as backbone models: \gls{mhrot} and \gls{ntl}. Our experiments are designed to test the hypothesis that \gls{loe} can mitigate the performance drop caused by training on contaminated image data. We experiment with three image datasets: CIFAR-10, Fashion-MNIST, and MVTEC \citep{bergmann2019mvtec}. These have been used in virtually all deep anomaly detection papers published at top-tier venues \citep{ruff2018deep,golan2018deep,hendrycks2019using,bergman2020classification,li2021cutpaste}, and we adopt these papers' experimental protocol here, as detailed below. 

\paragraph{Backbone models and baselines.}
We experiment with \gls{mhrot} and \gls{ntl}. In consistency with previous work \citep{hendrycks2019using}, we train \gls{mhrot} on raw images and \gls{ntl} on features outputted by an encoder pre-trained on ImageNet. We use the official code by the respective authors\footnote{\small\url{https://github.com/hendrycks/ss-ood.git}}\footnote{\small\url{https://github.com/boschresearch/NeuTraL-AD.git}}. \gls{ntl} is built upon the final pooling layer of a pre-trained ResNet152 for CIFAR-10 and F-MNIST (as suggested in \citet{defard2021padim}), and upon the third residual block of a pre-trained WideResNet50 for MVTEC (as suggested in \citet{reiss2021panda}).
Further implementation details of \gls{ntl} are in the \Cref{app:imp}.

Many existing baselines apply either blind updates or a refinement strategy to specific backbone models (see \Cref{sec:related}). However, a recent study showed that many of the classical anomaly detection methods such as autoencoders are no longer on par with modern self-supervised approaches \citep{alvarez2022revealing,hendrycks2019using} and in particular found NTL to perform best among 13 considered models.
For a more competitive and unified comparison with existing baselines in terms of the training strategy,
we hence adopt the two proposed \gls{loe} methods (\Cref{sec:method}) and the two baseline methods ``Blind'' and ``Refine'' (\Cref{sec:related}) to two backbone models.

\paragraph{Image datasets.}
On CIFAR-10 and F-MNIST, we follow the standard ``one-vs.-rest'' protocol of converting these data into anomaly detection datasets~\citep{ruff2018deep,golan2018deep,hendrycks2019using,bergman2020classification}. We create $C$ anomaly detection tasks (where $C$ is the number of classes), with each task considering one of the classes as normal and the union of all other classes as abnormal. For each task, 
the training set is a mixture of normal samples and a fraction of $\alpha_0$ abnormal samples.
For MVTEC, we use image features as the model inputs. The features are obtained from the third residual block of a WideResNet50 pre-trained on ImageNet as suggested in \citet{reiss2021panda}.
Since the MVTEC training set contains no anomalies, we contaminate it with artificial anomalies that we create by adding zero-mean Gaussian noise to the features of test set anomalies. We use a large variance for the additive noise (equal to the empirical variance of the anomalous features) to reduce information leakage from the test set into the training set.

\paragraph{Results.}
\begin{table}[t!]
	\caption{AUC ($\%$) with standard deviation for anomaly detection on CIFAR-10 and F-MNIST. For all experiments, we set the contamination ratio as $10\%$. \gls{loe} mitigates the performance drop when \gls{ntl} and \gls{mhrot} trained on the contaminated datasets.}
	\label{tab:image_results}
	\small
	\centering
	\vspace{2pt}
	\begin{tabular}{ll|cc}
        \hline
		     && CIFAR-10   & F-MNIST  \\
		\hline
        \multirow{4}{*}{\rotatebox{90}{NTL}} 
        &Blind &91.3$\pm$0.1 (-4.4) & 85.0$\pm$0.2 (-9.7) \\
        & Refine & 93.5$\pm$0.1 (-2.2) & 89.1$\pm$0.2 (-5.6) \\
        & LOE$_H$ (ours)& \textbf{94.9$\pm$0.2 (-0.8)} & \textbf{92.9$\pm$0.7 (-1.8)} \\
        & LOE$_S$ (ours)& \textbf{94.9$\pm$0.1 (-0.8)} & 92.5$\pm$0.1 (-2.2)\\
        \hline
        \multirow{4}{*}{\rotatebox{90}{MHRot}} 
        &Blind &84.0$\pm$0.5 (-4.2) &88.8$\pm$0.1 (-4.9)\\
        & Refine &84.4$\pm$0.1 (-3.8) &89.6$\pm$0.2 (-4.1)\\
        & LOE$_H$ (ours)&\textbf{86.4$\pm$0.5 (-1.8)} &\textbf{91.4$\pm$0.2 (-2.3)}\\
        & LOE$_S$ (ours)&86.3$\pm$0.2 (-1.9) &91.2$\pm$0.4 (-2.5)\\
        \hline
	\end{tabular}
\end{table}

\begin{table}[t!]
	\caption{AUC ($\%$) with standard deviation of \gls{ntl} for anomaly detection/segmentation on MVTEC. We set the contamination ratio of the training set as $10\%$ and $20\%$.}
	\label{tab:mvtec_results}
	\small
	\centering
	\vspace{2pt}
	\resizebox{\linewidth}{!}{
	\begin{tabular}{l|cc|cc}
        \hline
        &\multicolumn{2}{c|}{Detection} &\multicolumn{2}{c}{Segmentation}\\
        & $10\%$ & $20\%$ & $10\%$ & $20\%$\\ 
		\hline
        \multirow{2}{*}{Blind} & 94.2$\pm$0.5 & 89.4$\pm$0.3 & 96.17$\pm$0.08   & 95.09$\pm$0.17  \\
        & (-3.2) & (-8.0) & (-0.78)  & (-1.86)\\
        \multirow{2}{*}{Refine} & 95.3$\pm$0.5 & 93.2$\pm$0.3 & 96.55$\pm$0.04  & 96.09$\pm$0.06  \\
        & (-2.1) & (-4.2) & (-0.40) & (-0.86)\\
        LOE$_H$ & \textbf{95.9$\pm$0.9} & 92.9$\pm$0.4 & 95.97$\pm$0.22  & 93.29$\pm$0.21  \\
        (ours)& (-1.5) & (-4.5) & (-0.98) & (-3.66)\\
        LOE$_S$ & 95.4$\pm$0.5  & \textbf{93.6$\pm$0.3} & \textbf{96.56$\pm$0.04}   & \textbf{96.11$\pm$0.05}  \\
        (ours)& (-2.0) & (-3.8) & (-0.39) & (-0.84)\\
        \hline
	\end{tabular}
	}
\end{table}

\begin{figure*}[t!]
    \centering
    	\begin{subfigure}[b]{0.6\linewidth}
		\includegraphics[width=\linewidth]{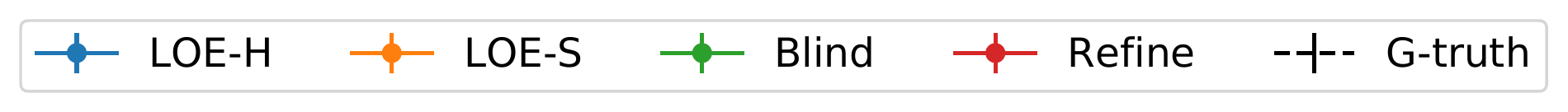}
	\end{subfigure}	\\
	\begin{subfigure}[b]{0.245\linewidth}
		\includegraphics[width=\linewidth]{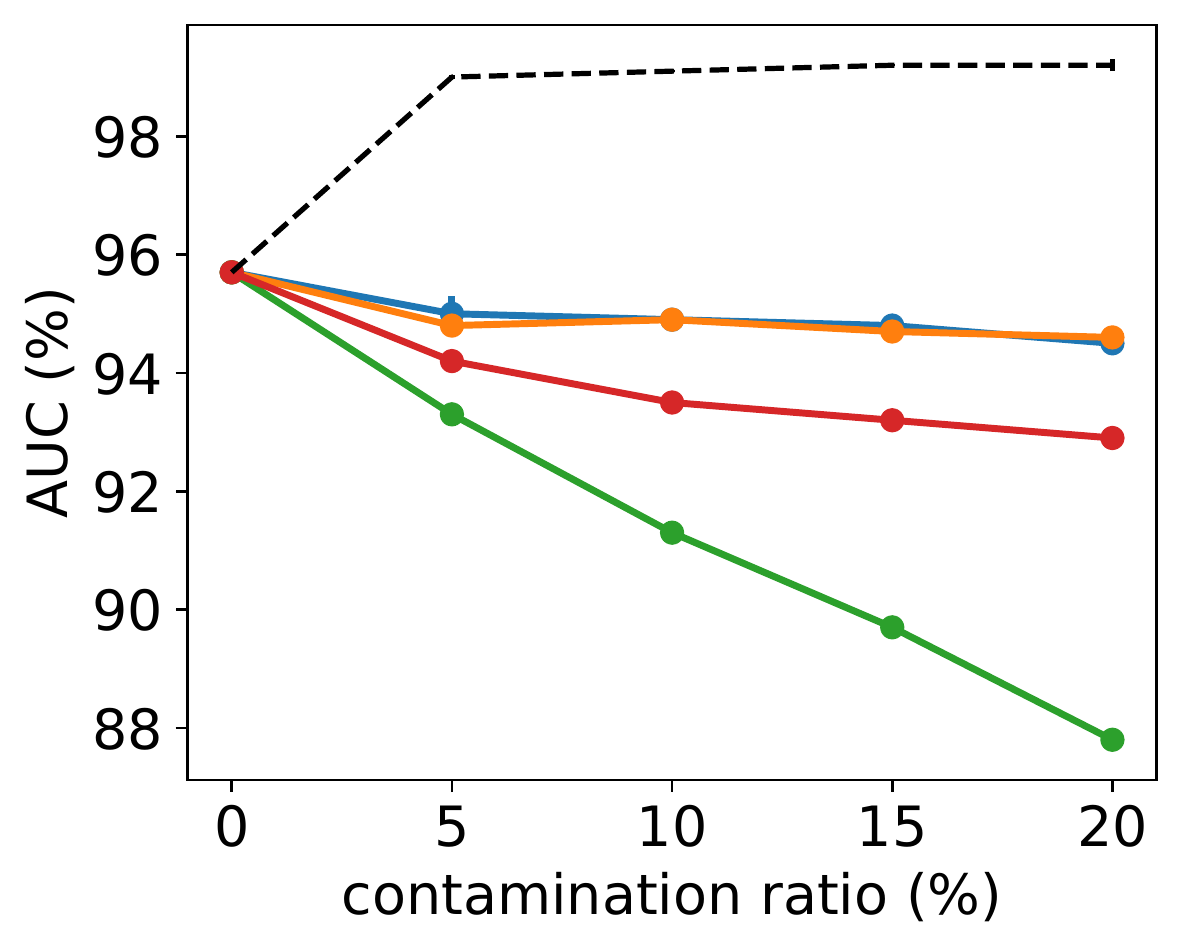}
	\caption{CIFAR-10}
	\end{subfigure}
		\begin{subfigure}[b]{0.245\linewidth}
		\includegraphics[width=\linewidth]{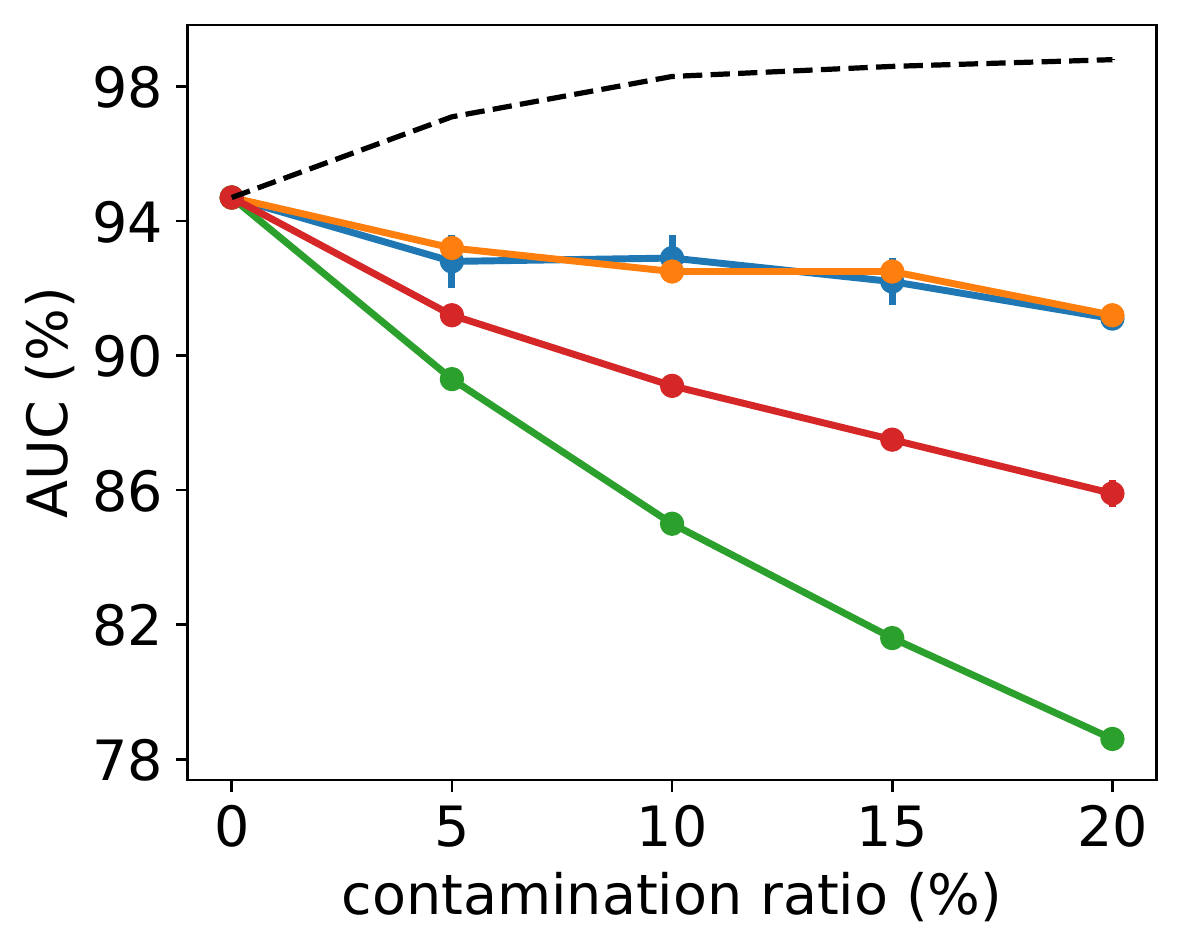}
	\caption{F-MNIST}
	\end{subfigure}
	\begin{subfigure}[b]{0.245\linewidth}
		\includegraphics[width=\linewidth]{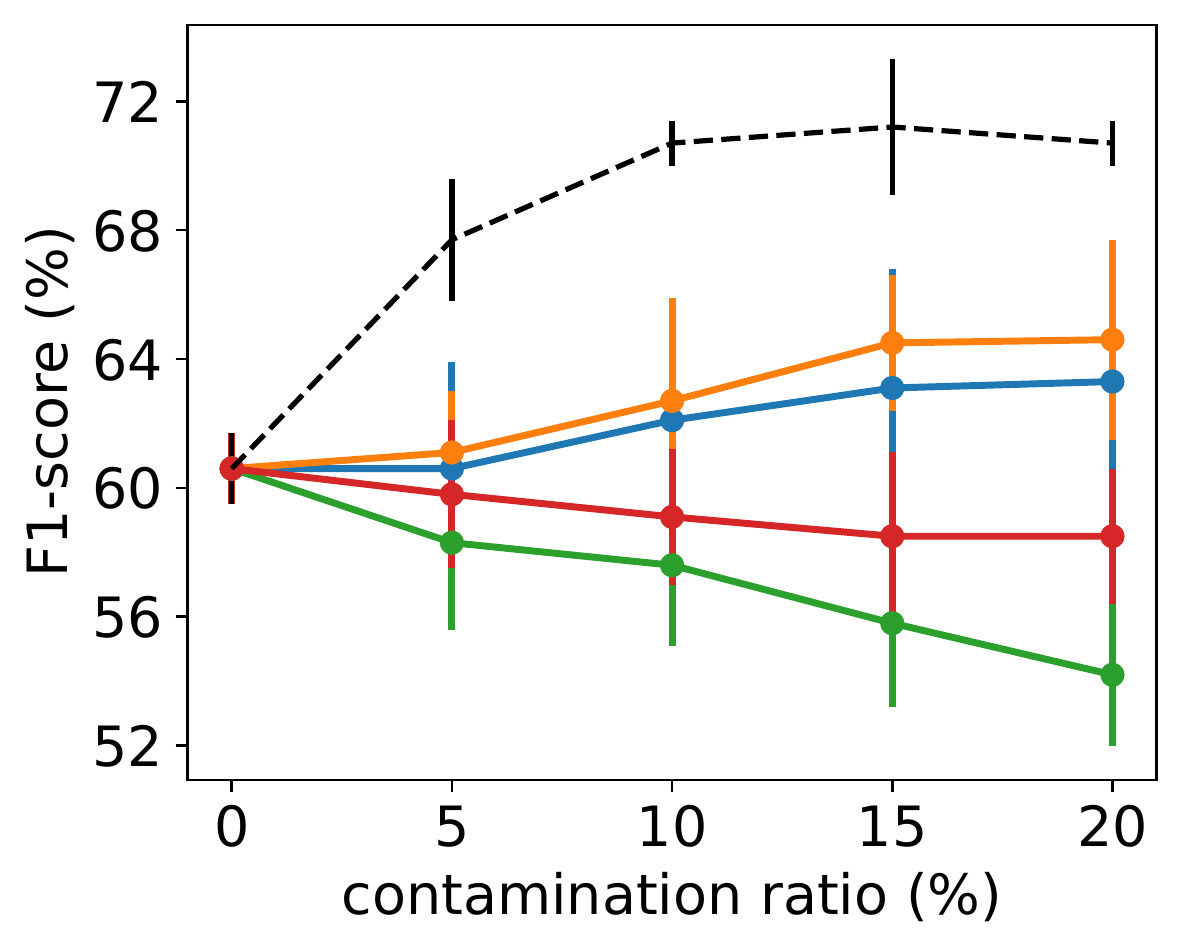}
    \caption{Arrhythmia}
	\end{subfigure}
	\begin{subfigure}[b]{0.245\linewidth}
		\includegraphics[width=\linewidth]{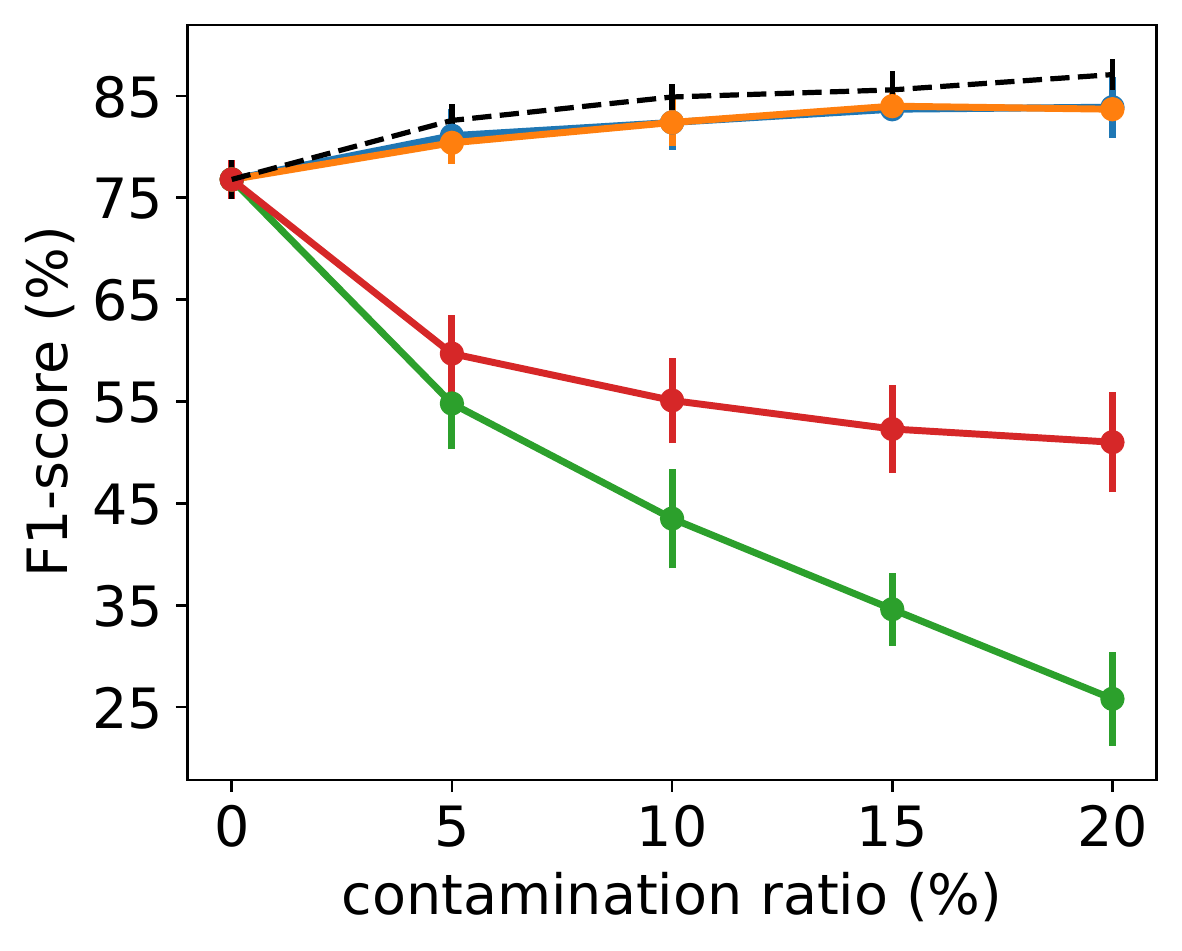}
    \caption{Thyroid}
	\end{subfigure}\\
    \caption{Anomaly detection performance of \gls{ntl} on CIFAR-10, F-MNIST, and two tabular datasets (Arrhythmia and Thyroid) with  $\alpha_0 \in \{5\%, 10\%, 15\%, 20\%\}$. \gls{loe} (ours) consistently outperforms the ``Blind'' and ``Refine'' on various contamination ratios.}
    \label{fig:acc_ratio}
\end{figure*}
\begin{table*}[t!]
	\caption{F1-score ($\%$) for anomaly detection on $30$ tabular datasets studied in \citep{shenkar2022anomaly}. We set $\alpha_0=\alpha=10\%$ in all experiments. 
	\gls{loe} (proposed) outperforms the ``Blind'' and ``Refine'' consistently. (See \cref{tab:tabular_f1,tab:tabular_auc} for more details, including AUCs.)}
	\label{tab:tabular_small}
	\small
	\centering
	\vspace{2pt}
	\resizebox{\linewidth}{!}{
	\begin{tabular}{l|cccc|cccc}
        \hline
        &\multicolumn{4}{c|}{NTL}&\multicolumn{4}{c}{ICL} \\
        &Blind& Refine& LOE$_H$ (ours)& LOE$_S$ (ours)&Blind& Refine& LOE$_H$ (ours)& LOE$_S$ (ours)\\
        \hline
abalone	& 37.9$\pm$13.4& 55.2$\pm$15.9& 42.8$\pm$26.9& \textbf{59.3$\pm$12.0}& 50.9$\pm$1.5& \textbf{54.3$\pm$2.9}& 53.4$\pm$5.2& 51.7$\pm$2.4\\ 
annthyroid	& 29.7$\pm$3.5& 42.7$\pm$7.1& 47.7$\pm$11.4& \textbf{50.3$\pm$4.5}& 29.1$\pm$2.2& 38.5$\pm$2.1& \textbf{48.7$\pm$7.6}& 43.0$\pm$8.8\\ 
arrhythmia	& 57.6$\pm$2.5& 59.1$\pm$2.1& 62.1$\pm$2.8& \textbf{62.7$\pm$3.3}& 53.9$\pm$0.7& 60.9$\pm$2.2& 62.4$\pm$1.8& \textbf{63.6$\pm$2.1}\\ 
breastw	& 84.0$\pm$1.8& 93.1$\pm$0.9& \textbf{95.6$\pm$0.4}& 95.3$\pm$0.4& 92.6$\pm$1.1& 93.4$\pm$1.0& \textbf{96.0$\pm$0.6}& 95.7$\pm$0.6\\ 
cardio	& 21.8$\pm$4.9& 45.2$\pm$7.9& \textbf{73.0$\pm$7.9}& 57.8$\pm$5.5& 50.2$\pm$4.5& 56.2$\pm$3.4& \textbf{71.1$\pm$3.2}& 62.2$\pm$2.7\\ 
ecoli	& 0.0$\pm$0.0& 88.9$\pm$14.1& \textbf{100$\pm$0.0}& \textbf{100$\pm$0.0}& 17.8$\pm$15.1& 46.7$\pm$25.7& \textbf{75.6$\pm$4.4}& \textbf{75.6$\pm$4.4}\\ 
forest cover	& 20.4$\pm$4.0& 56.2$\pm$4.9& 61.1$\pm$34.9& \textbf{67.6$\pm$30.6}& 9.2$\pm$4.5& 8.0$\pm$3.6& 6.8$\pm$3.6& \textbf{11.1$\pm$2.1}\\ 
glass	& 11.1$\pm$7.0& 15.6$\pm$5.4& 17.8$\pm$5.4& \textbf{20.0$\pm$8.3}& 8.9$\pm$4.4& \textbf{11.1$\pm$0.0}& \textbf{11.1$\pm$7.0}& 8.9$\pm$8.3\\ 
ionosphere	& 89.0$\pm$1.5& 91.0$\pm$2.0& 91.0$\pm$1.7& \textbf{91.3$\pm$2.2}& 86.5$\pm$1.1& 85.9$\pm$2.3& 85.7$\pm$2.8& \textbf{88.6$\pm$0.6}\\ 
kdd	& 95.9$\pm$0.0& 96.0$\pm$1.1& 98.1$\pm$0.4& \textbf{98.4$\pm$0.1}& 99.3$\pm$0.1& 99.4$\pm$0.1& \textbf{99.5$\pm$0.0}& 99.4$\pm$0.0\\ 
kddrev	& 98.4$\pm$0.1& 98.4$\pm$0.2& 89.1$\pm$1.7& \textbf{98.6$\pm$0.0}& 97.9$\pm$0.5& 98.4$\pm$0.4& \textbf{98.8$\pm$0.1}& 98.2$\pm$0.4\\ 
letter	& 36.4$\pm$3.6& 44.4$\pm$3.1& 25.4$\pm$10.0& \textbf{45.6$\pm$10.6}& 43.0$\pm$2.5& 51.2$\pm$3.7& \textbf{54.4$\pm$5.6}& 47.2$\pm$4.9\\ 
lympho	& 53.3$\pm$12.5& 60.0$\pm$8.2& 60.0$\pm$13.3& \textbf{73.3$\pm$22.6}& 43.3$\pm$8.2& 60.0$\pm$8.2& 80.0$\pm$12.5& \textbf{83.3$\pm$10.5}\\ 
mammogra.	& 5.5$\pm$2.8& 2.6$\pm$1.7& 3.3$\pm$1.6& \textbf{13.5$\pm$3.8}& 8.8$\pm$1.9& 11.4$\pm$1.9& 34.0$\pm$20.2& \textbf{42.8$\pm$17.6}\\ 
mnist tabular	& 78.6$\pm$0.5& \textbf{80.3$\pm$1.1}& 71.8$\pm$1.8& 76.3$\pm$2.1& 72.1$\pm$1.0& 80.7$\pm$0.7& \textbf{86.0$\pm$0.4}& 79.2$\pm$0.9\\ 
mulcross	& 45.5$\pm$9.6& \textbf{58.2$\pm$3.5}& \textbf{58.2$\pm$6.2}& 50.1$\pm$8.9& 70.4$\pm$13.4& 94.4$\pm$6.3& \textbf{100$\pm$0.0}& 99.9$\pm$0.1\\ 
musk	& 21.0$\pm$3.3& 98.8$\pm$0.4& \textbf{100$\pm$0.0}& \textbf{100$\pm$0.0}& 6.2$\pm$3.0& \textbf{100$\pm$0.0}& \textbf{100$\pm$0.0}& \textbf{100$\pm$0.0}\\ 
optdigits	& 0.2$\pm$0.3& 1.5$\pm$0.3& 41.7$\pm$45.9& \textbf{59.1$\pm$48.2}& 0.8$\pm$0.5& \textbf{1.3$\pm$1.1}& 1.2$\pm$1.0& 0.9$\pm$0.5\\ 
pendigits	& 5.0$\pm$2.5& 32.6$\pm$10.0& 79.4$\pm$4.7& \textbf{81.9$\pm$4.3}& 10.3$\pm$4.6& 30.1$\pm$8.5& 80.3$\pm$6.1& \textbf{88.6$\pm$2.2}\\ 
pima	& 60.3$\pm$2.6& 61.0$\pm$1.9& \textbf{61.3$\pm$2.4}& 61.0$\pm$0.9& 58.1$\pm$2.9& 59.3$\pm$1.4& \textbf{63.0$\pm$1.0}& 60.1$\pm$1.4\\ 
satellite	& 73.6$\pm$0.4& 74.1$\pm$0.3& \textbf{74.8$\pm$0.4}& 74.7$\pm$0.1& 72.7$\pm$1.3& 72.7$\pm$0.6& \textbf{73.6$\pm$0.2}& 73.2$\pm$0.6\\ 
satimage	& 26.8$\pm$1.5& 86.8$\pm$4.0& 90.7$\pm$1.1& \textbf{91.0$\pm$0.7}& 7.3$\pm$0.6& 85.1$\pm$1.4& 91.3$\pm$1.1& \textbf{91.5$\pm$0.9}\\ 
seismic	& 11.9$\pm$1.8& 11.5$\pm$1.0& \textbf{18.1$\pm$0.7}& 17.1$\pm$0.6& 14.9$\pm$1.4& 17.3$\pm$2.1& 23.6$\pm$2.8& \textbf{24.2$\pm$1.4}\\ 
shuttle	& 97.0$\pm$0.3& 97.0$\pm$0.2& \textbf{97.1$\pm$0.2}& 97.0$\pm$0.2& 96.6$\pm$0.2& 96.7$\pm$0.1& 96.9$\pm$0.1& \textbf{97.0$\pm$0.2}\\ 
speech	& 6.9$\pm$1.2& 8.2$\pm$2.1& 43.3$\pm$5.6& \textbf{50.8$\pm$2.5}& 0.3$\pm$0.7& 1.6$\pm$1.0& \textbf{2.0$\pm$0.7}& 0.7$\pm$0.8\\ 
thyroid	& 43.4$\pm$5.5& 55.1$\pm$4.2& \textbf{82.4$\pm$2.7}& \textbf{82.4$\pm$2.3}& 45.8$\pm$7.3& 71.6$\pm$2.4& \textbf{83.2$\pm$2.9}& 80.9$\pm$2.5\\ 
vertebral	& 22.0$\pm$4.5& 21.3$\pm$4.5& 22.7$\pm$11.0& \textbf{25.3$\pm$4.0}& 8.9$\pm$3.1& 8.9$\pm$4.2& 7.8$\pm$4.2& \textbf{10.0$\pm$2.7}\\ 
vowels	& 36.0$\pm$1.8& 50.4$\pm$8.8& \textbf{62.8$\pm$9.5}& 48.4$\pm$6.6& 42.1$\pm$9.0& 60.4$\pm$7.9& \textbf{81.6$\pm$2.9}& 74.4$\pm$8.0\\ 
wbc	& 25.7$\pm$12.3& 45.7$\pm$15.5& \textbf{76.2$\pm$6.0}& 69.5$\pm$3.8& 50.5$\pm$5.7& 50.5$\pm$2.3& \textbf{61.0$\pm$4.7}& \textbf{61.0$\pm$1.9}\\ 
wine	& 24.0$\pm$18.5& 66.0$\pm$12.0& 90.0$\pm$0.0& \textbf{92.0$\pm$4.0}& 4.0$\pm$4.9& 10.0$\pm$8.9& 98.0$\pm$4.0& \textbf{100$\pm$0.0}\\ 
\hline
	\end{tabular}
	}
\end{table*}

We present the experimental results of CIFAR-10 and F-MNIST in \cref{tab:image_results}, where we set the contamination ratio $\alpha_0=\alpha=0.1$. 
The results are reported as the mean and standard deviation of three runs with different model initialization and anomaly samples for the contamination.
The number in the brackets is the average performance difference from the model trained on clean data. 
Our proposed methods consistently outperform the baselines and mitigate the performance drop between the model trained on clean data vs. the same model trained on contaminated data. Specifically, with \gls{ntl}, \gls{loe} significantly improves over the best-performing baseline, ``Refine'', by $1.4\%$ and $3.8\%$ AUC on CIFAR-10 and F-MNIST, respectively. On CIFAR-10, our methods have only $0.8\%$ AUC lower than when training on the normal dataset. When we use another state-of-the-art method \gls{mhrot} on raw images, our \gls{loe} methods outperform the baselines by about $2\%$ AUC on both datasets. 

We also evaluate our methods with \gls{ntl} at various contamination ratios (from $5\%$ to $20\%$) in \cref{fig:acc_ratio} (a) and (b). We can see 1) adding labeled anomalies (G-truth) boosts performance, and 2) among all methods that do not have ground truth labels, the proposed \gls{loe} methods achieve the best performance consistently at all contamination ratios. 


We also experimented on anomaly detection and segmentation on the MVTEC dataset. Results are shown in \cref{tab:mvtec_results}, where we evaluated the methods on two contamination ratios ($10\%$ and $20\%$). Our method improves over the ``Blind'' and ``Refine'' baselines in all experimental settings. 


\subsection{Experiments on Tabular Data}
Tabular data is another important application area of anomaly detection. Many data sets in the healthcare and cybersecurity domains are tabular.
Our empirical study demonstrates that \gls{loe} yields the best performance for two popular backbone models on a comprehensive set of contaminated tabular datasets. 

\paragraph{Tabular datasets.}
We study all 30 tabular datasets used in the empirical analysis of a recent state-of-the-art paper \citep{shenkar2022anomaly}. These include the frequently-studied small-scale Arrhythmia and Thyroid medical datasets, the large-scale cyber intrusion detection datasets KDD and KDDRev, and multi-dimensional point datasets from the outlier detection datasets\footnote{\url{http://odds.cs.stonybrook.edu/}}. We follow the pre-processing and train-test split of the datasets in \citet{shenkar2022anomaly}. To corrupt the training set, we create artificial anomalies by adding zero-mean Gaussian noise to anomalies from the test set. 
We use a large variance for the additive noise (equal to the empirical variance of the anomalies in the test set) to reduce information leakage from the test set into the training set.

\paragraph{Backbone models and baselines.}
We consider two advanced deep anomaly detection methods for tabular data described in \Cref{sec:examples}: \gls{ntl} and \gls{icl}. For \gls{ntl}, we use nine transformations and multi-layer perceptrons for neural transformations and the encoder on all datasets. Further details are provided in \Cref{app:imp}. 
For \gls{icl}, we use the code provided by the authors.
We implement the proposed \gls{loe} methods (\Cref{sec:method}) and the ``Blind" and ``Refine" baselines (\Cref{sec:related}) with both backbone models. 

\paragraph{Results.}
We report F1-scores for 30 tabular datasets in \cref{tab:tabular_small}. The results are reported as the mean and standard derivation of five runs with different model initializations and random training set split. We set the contamination ratio $\alpha_0=\alpha=0.1$ for all datasets. 
More detailed results, including AUCs and the performance degradation over clean data, are provided in \Cref{app:results} (\cref{tab:tabular_f1,tab:tabular_auc}). 

\gls{loe} outperforms the ``Blind" and ``Refine" baselines consistently. Remarkably, on some datasets, \gls{loe} trained on contaminated data can achieve better results than on clean data (as shown in \cref{tab:tabular_f1}), suggesting that the latent anomalies provide a positive learning signal.  
This effect can be seen when increasing the contamination ratio on the Arrhythmia and Thyroid datasets (\cref{fig:acc_ratio} (c) and (d)). \citet{hendrycks2018deep} noticed a similar phenomenon when adding \emph{labeled} auxiliary outliers; these known anomalies help the model learn better region boundaries for normal data. Our results suggest that even \emph{unlabelled} anomalies, when properly inferred, can improve the performance of an anomaly detector.
Overall, we conclude that \gls{loe} significantly improves the performance of anomaly detection methods on contaminated tabular datasets.

\begin{table}[t!]
	\caption{AUC ($\%$) for different contamination ratios for a video frame anomaly detection benchmark proposed in \citep{pang2020self}. LOE$_S$ (proposed) achieves state-of-the-art performance.}
	\label{tab:video_results}
	\centering
	\vspace{2pt}
 	\resizebox{\linewidth}{!}{
	\begin{tabular}{l|ccc}
        \hline
        Method    &\multicolumn{3}{c}{Contamination Ratio} \\
		     & 10\%   & 20\%  & 30\%$^*$ \\
		\hline
		\citep{tudor2017unmasking} &- &- &68.4 \\
		\citep{liu2018classifier} &- &- &69.0 \\
		\citep{del2016discriminative} &- &- &59.6 \\
		\citep{sugiyama2013rapid} &55.0 &56.0 &56.3 \\
		\citep{pang2020self} &68.0 &70.0 &\textbf{71.7} \\
        Blind &85.2$\pm$1.0 &76.0$\pm$2.7 &66.6$\pm$2.6 \\
        Refine & 82.7$\pm$1.5 &74.9$\pm$2.4 &69.3$\pm$0.7 \\
        LOE$_H$ (ours) & 82.3$\pm$1.6 &59.6$\pm$3.8 &56.8$\pm$9.5 \\
        LOE$_S$ (ours)& \textbf{86.8$\pm$1.2} &\textbf{79.2$\pm$1.3} &\textbf{71.5$\pm$2.4} \\
        \hline
        \multicolumn{4}{l}{{ 
         $^*$Default setup in \citep{pang2020self}, corresponding to $\alpha_0\approx 30 \%$.
        }}\\
	\end{tabular}
 	}
\end{table}
\begin{figure*}[t]
    \centering
	\begin{subfigure}[b]{0.28\linewidth}
		\includegraphics[width=\linewidth]{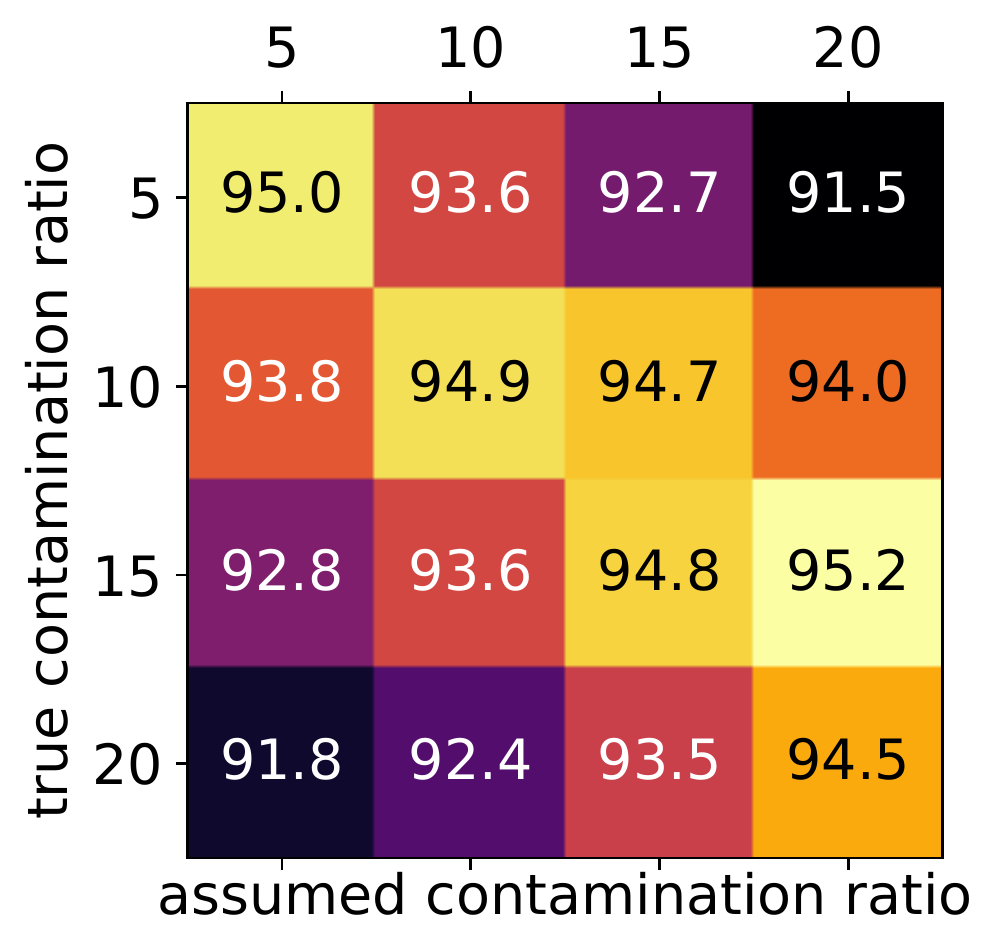}
	\caption{LOE$_H$}
	\end{subfigure}
	\begin{subfigure}[b]{0.28\linewidth}
		\includegraphics[width=\linewidth]{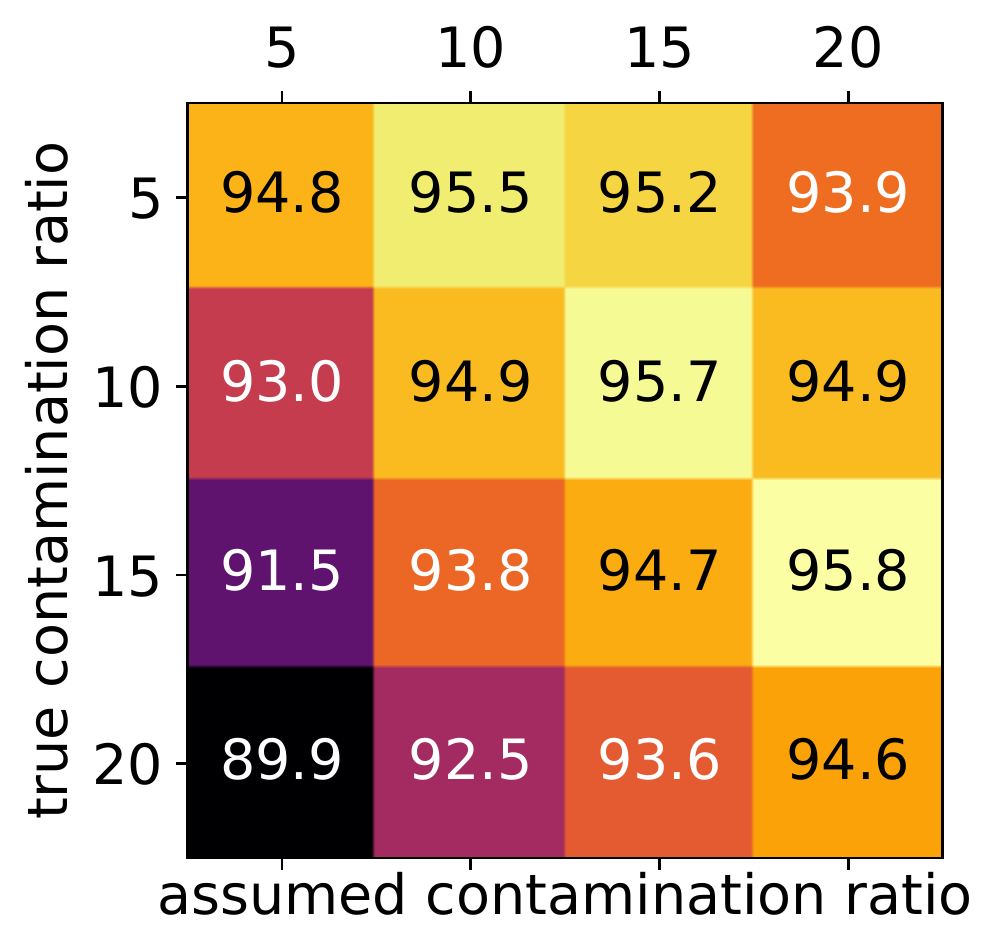}
    \caption{LOE$_S$}
	\end{subfigure}
		\begin{subfigure}[b]{0.28\linewidth}
		\includegraphics[width=\linewidth]{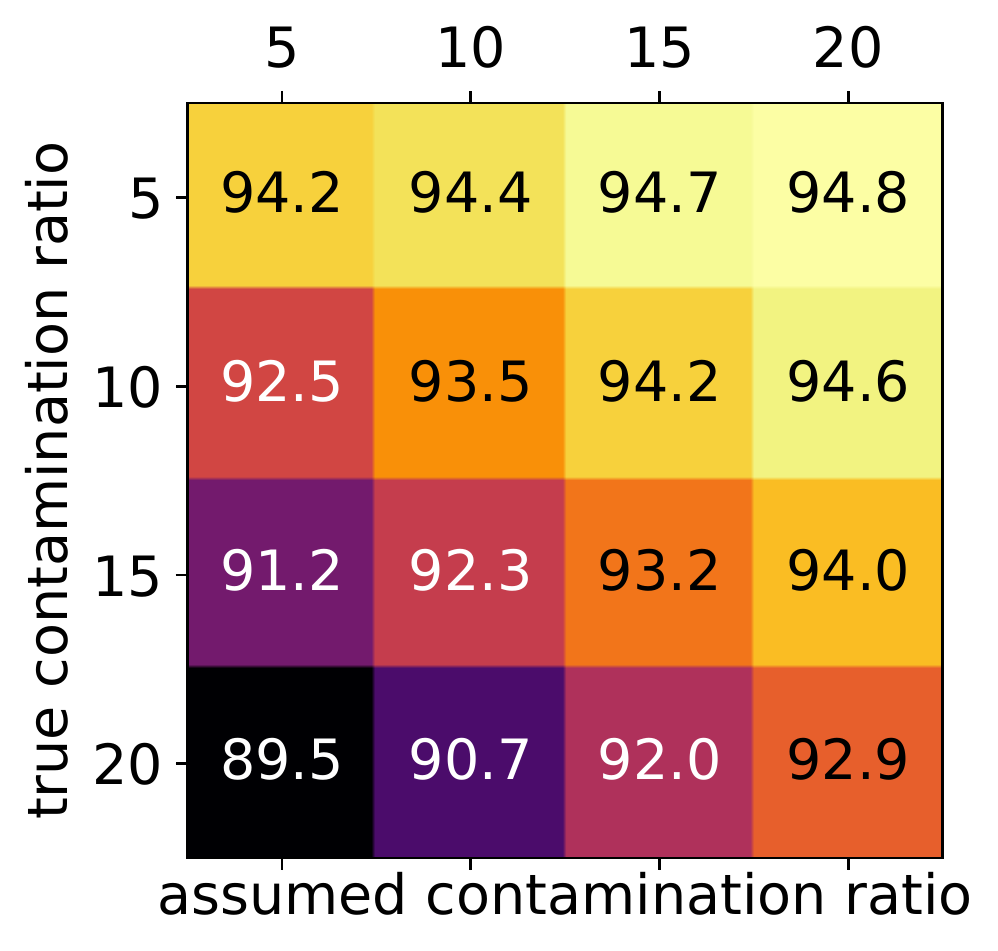}
    \caption{Refine}
	\end{subfigure}
    \caption{A sensitivity study of the robustness of \gls{loe}$_H$, \gls{loe}$_S$, and ``Refine'' to the mis-specified contamination ratio. We evaluate them with \gls{ntl} on CIFAR-10 in terms of AUC. \gls{loe}$_H$ and \gls{loe}$_S$ yield robust results and outperform ``Refine'' in the most cases. }
    \label{fig:ablation}
\end{figure*}

\subsection{Experiments on Video Data}
In addition to image and tabular data, we also evaluate our methods on a video frame anomaly detection benchmark also studied in \citep{pang2020self}. The goal is to identify video frames that contain unusual objects or abnormal events. 
Experiments show that our methods achieve state-of-the-art performance on this benchmark.

\paragraph{Video dataset.}
We study UCSD Peds1\footnote{
  \url{http://www.svcl.ucsd.edu/projects/anomaly/dataset.htm}
}, a popular benchmark for video anomaly detection. It contains surveillance videos of a pedestrian walkway. Non-pedestrian and unusual behavior is labeled as abnormal. 
The data set contains 34 training video clips and 36 testing video clips, where all frames in the training set are normal and about half of the testing frames are abnormal. We follow the data preprocessing protocol of \citet{pang2020self} for dividing the data into training and test sets. To realize different contamination ratios, we randomly remove some abnormal frames from the training set but the test set is fixed.


\paragraph{Backbone models and baselines.}
In addition to the ``Blind" and ``Refine'' baselines, we compare to ~\citep{pang2020self} (a ranking-based state-of-the-art method for video frame anomaly detection already described in \cref{sec:related}) and all baselines reported in that paper~\citep{sugiyama2013rapid,liu2012isolation, del2016discriminative, tudor2017unmasking, liu2018classifier}. 


We implement the proposed \gls{loe} methods, the ``Blind'', and the ``Refine'' baselines with \gls{ntl} as the backbone model. We use a pre-trained ResNet50 on ImageNet as a feature extractor, whose output is then sent into an \gls{ntl}. The feature extractor and \gls{ntl} are jointly optimized during training.

\paragraph{Results.}
We report the results in \cref{tab:video_results}. Our soft \gls{loe} method achieves the best performance across different contamination ratios. Our method outperforms Deep Ordinal Regression \citep{pang2020self} by 18.8\% and 9.2\% AUC on the contamination ratios of 10\% and 20\%, respectively. \gls{loe}$_S$ outperforms the ``Blind" and ``Refine" baselines significantly on various contamination ratios.

\subsection{Sensitivity Study}
The hyperparameter $\alpha$ characterizes the assumed fraction of anomalies in our training data. 
Here, we evaluate its robustness under different ground truth contamination ratios. 
We run \gls{loe}$_H$ and \gls{loe}$_S$ with \gls{ntl} on CIFAR-10 with varying true anomaly ratios $\alpha_0$ and different hyperparameters~$\alpha$. 
We present the results in a matrix accommodating the two variables. The diagonal values report the results when correctly setting the contamination ratio.

\gls{loe}$_H$ (\cref{fig:ablation} (a)) shows considerable robustness: the method suffers at most $1.4\%$ performance degradation when the hyperparameter $\alpha$ is off by $5\%$, and is always better than ``Blind". It always outperforms ``Refine" (\cref{fig:ablation} (c)) when erroneously setting a smaller $\alpha$ than the true ratio $\alpha_0$.  
\gls{loe}$_S$ (\cref{fig:ablation} (b)) also shows robustness, especially when erroneously setting a larger $\alpha$ than $\alpha_0$. The method is always better than ``Refine" (\cref{fig:ablation} (c)) when the hyperparameter $\alpha$ is off by up to $15\%$, and always outperforms ``Blind".


\section{Conclusion}
\label{sec:conclusion}
We propose Latent Outlier Exposure (\gls{loe}): a domain-independent approach for training anomaly detectors on a dataset contaminated by unidentified anomalies. During training, \gls{loe} jointly infers anomalous data in the training set while updating its parameters by solving a mixed continuous-discrete optimization problem; iteratively updating the model and its predicted anomalies. Similar to outlier exposure \citep{hendrycks2018deep}, \gls{loe} extracts a learning signal from both normal and abnormal samples by considering a combination of two losses for both normal and (assumed) abnormal data, respectively. Our approach can be applied to a variety of anomaly detection benchmarks and loss functions. As demonstrated in our comprehensive empirical study, \gls{loe} yields significant performance improvements on all three of image, tabular, and video data. 
\section*{Acknowledgements}
Stephan Mandt acknowledges support by the National Science Foundation (NSF) under the NSF CAREER Award 2047418; NSF Grants 1928718, 2003237 and 2007719; the Department of Energy under grant DE-SC0022331, as well as gifts from Intel, Disney, and Qualcomm. This material is in part based upon work supported by the Defense Advanced Research Projects Agency (DARPA) under Contract No. HR001120C0021. Any opinions, findings and conclusions or recommendations expressed in this material are those of the author(s) and do not necessarily reflect the views of DARPA. 
Marius Kloft acknowledges support by the Carl-Zeiss Foundation, the DFG awards KL 2698/2-1 and KL 2698/5-1, and the BMBF awards 01$|$S18051A, 03$|$B0770E, and 01$|$S21010C.
We thank Sam Showalter for providing helpful feedback on our manuscript.

The Bosch Group is carbon neutral. Administration, manufacturing and research activities do no longer leave a carbon footprint. This also includes GPU clusters on which the experiments have been performed.

\bibliography{ref}
\bibliographystyle{icml2022}

\clearpage
\appendix
\section{Details on Toy Data Experiments}
\label{app:toy}
We generate the toy data with a three-component Gaussian mixture. The normal data is generated from $p_n = {\cal N}(\x; [1,1], 0.07I)$, and the anomalies are sampled from $p_{a}={\cal N}(\x; [-0.25, 2.5], 0.03I) + {\cal N}(\x; [-1., 0.5], 0.03I)$. There are 90 normal samples and 10 abnormal samples. All samples are mixed up as the contaminated training set.

To learn a anomaly detector, we used one-class Deep SVDD~\citep{ruff2018deep} to train a one-layer radial basis function (RBF) network where the Gaussian function is used as the RBF. The hidden layer contains three neurons whose centers are fixed at the center of each component and whose scales are optimized during training. The output of the RBF net is a linear combination of the outputs of hidden layers. Here we set the model output to be a 1D scalar, as the projected data representation of Deep SVDD. 

For Deep SVDD configuration, we randomly initialized the model center (not to be confused with the center of the Gaussian RBF) and made it learnable during training. We also added the bias term in the last layer. Although setting a learnable center and adding bias terms are not recommended for Deep SVDD~\citep{ruff2018deep} due to the all-zero trivial solution, we found these practices make the model flexible and converge well and learn a much better anomaly detector than vice verse, probably because the random initialization and small learning rate serve as regularization and the model converges to a local optimum before collapses to the trivial solution. During training, we used Adam~\citep{kingma2014adam} stochastic optimizer and set the mini-batch size to be 25. The learning rate is $0.01$, and we trained the model for 200 epochs. The decision boundary in \Cref{fig:toy_data} plots the 90\% fraction of the anomaly scores.

\section{Baseline Details}
\label{app:baseline}
Across all experiments, we employ two baselines that do not utilize anomalies to help training the models. The baselines are either completely blind to anomalies, or drop the perceived anomalies' information. Normally training a model without recognizing anomalies serves as our first baseline. Since this baseline doesn't take any actions to the anomalies in the contaminated training data and is actually blind to the anomalies that exist, we name it \textit{Blind}. Mathematically, Blind sets $y_i=0$ in Eq.~\ref{eq:loe-loss} for all samples. 

The second baseline filters out anomalies and refines the training data: at every mini-batch update, it first ranks the mini-batch data according to the anomaly scores given current detection model, then removes top $\alpha$ most likely anomalous samples from the mini-batch. The remaining samples performs the model update. We name the second baseline \textit{Refine}, which still follows Alg.~\ref{alg:loe} but removes $\La$ in Eq.~\ref{eq:loe-loss}. Both these two baselines take limited actions to the anomalies. We use them to contrast our proposed methods and highlight the useful information contained in unseen anomalies.

\section{Implementation Details}
\label{app:imp}
We apply \gls{ntl} to all datasets including both visual datasets and tabular datasets. Below we provide the implementation details of \gls{ntl} on each class of datasets.
\paragraph{\gls{ntl} on image data}
\gls{ntl} is built upon the final pooling layer of a pre-trained ResNet152 on CIFAR-10 and F-MNIST (as suggested in \citet{defard2021padim}), and upon the third residual block of a pre-trained WideResNet50 on MVTEC (as suggested in \citet{reiss2021panda}). On all image datasets, the pre-trained feature extractors are frozen during training. We set the number of transformations as 15 and use three linear layers with  intermediate 1d batchnorm layers and ReLU activations for transformations modelling. The hidden sizes of the transformation networks are $[2048,2048,2048]$ on CIFAR-10 and F-MNIST, and $[1024,1024,1024]$ on MVTEC. The encoder is one linear layer with units of $256$ for CIFAR-10 and MVTEC, and is two linear layers of size $[1024,256]$ with an intermediate ReLU activation for F-MNIST. On CIFAR-10, we set mini-batch size to be 500, learning rate to be 4e-4, 30 training epochs with Adam optimizer. On F-MNIST, we set mini-batch size to be 500, learning rate to be 2e-4, 30 training epochs with Adam optimizer. On MVTEC, we set mini-batch size to be 40, learning rate to be 2e-4, 30 training epochs with Adam optimizer.
For the ``Refine'' baseline and our methods we set the number of warm-up epochs as two on all image datasets.

\paragraph{\gls{ntl} on tabular data}
On all tabular data, we set the number of transformations to 9, use two fully-connected network layers for the transformations and four fully-connected network layers for the encoder. The hidden size of layers in the transformation networks and the encoder is two times the data dimension for low dimensional data, and 64 for high dimensional data. The embedding size is two times the data dimension for low dimensional data, and 32 for high dimensional data.
The transformations are either parametrized as the transformation network directly or a residual connection of the transformation network and the original sample. We search the best-performed transformation parameterization and other hyperparameters based on the performance of the model trained on clean data. 
We use Adam optimizer with a learning rate chosen from $[5e-4, 1e-3, 2e-3]$. 
For the ``Refine'' baseline and our methods we set the number of warm-up epochs as two for small datasets and as one for large datasets.

\paragraph{\gls{ntl} on video data}
Following the suggestions of \citet{pang2020self}, we first extract frame features through a ResNet50 pretrained on ImageNet. The features are sent to an NTL with the same backbone model as used on CIFAR-10 (see NTL on image data) except that 9 transformations are used. Both the ResNet50 and NTL are updated from end to end. During training, we use Adam stochastic optimizer with the batch size set to be 192 and learning rate set 1e-4. We update the model for 3 epochs and report the results with three independent runs.

\paragraph{\gls{mhrot} on image data}
\gls{mhrot} \citep{hendrycks2019using} applies self-supervised learning on hand-crafted image transformations including rotation, horizontal shift, and vertical shift. The learner learns to solve three different tasks: one for predicting rotation ($r\in \mathcal{R}\equiv\{0^{\circ}, \pm90^{\circ}, 180^{\circ}\}$), one for predicting vertical shift ($s^v \in \mathcal{S}^v\equiv\{0\text{\,px}, \pm 8\text{\,px}\}$), and one for predicting horizontal shift ($s^h \in \mathcal{S}^h\equiv\{0\text{\,px}, \pm 8\text{\,px}\}$). 
We define the composition of rotation, vertical shift, and horizontal shift as $T \in {\cal T}\equiv\{r\circ s^v \circ s^h \mid r\in \mathcal{R}, s^v \in \mathcal{S}^v, s^h \in \mathcal{S}^h\}$. We also define the head labels $t^1_k=r_a, t^2_k=s^v_b, t^3_k=s^h_c$ for a specific composed transformation $T_k=r_a\circ s^v_b \circ s^h_c$.
Overall, there are 36 transformations.

We implement the model on the top of GOAD~\citep{bergman2020classification}, a similar self-supervised anomaly detector. The backbone model is a WideResNet16-4. Anomaly scores is used for ranking in the mini-batch in pseudo label assignments. For F-MNIST, we use $\Ln$, the normality training loss, as the anomaly score. For CIFAR-10, we find that using a separate anomaly score mentioned in \citep{bergman2020classification} leads to much better results than the original training loss anomaly score. 

During training, we set mini-batch size to be 10, learning rate to be 1e-3 for CIFAR-10 and 1e-4 for F-MNIST, 16 training epochs for CIFAR-10 and 3 training epochs for F-MNIST with Adam optimizer. We report the results with 3-5 independent runs.

\section{Additional Experimental Results}
\label{app:results}
We provide additional results of the experiments on tabular datasets. We report the F1-scores in Table~\ref{tab:tabular_f1} and the AUCs in Table~\ref{tab:tabular_auc}. The number in the brackets is the average performance difference from the model trained on clean data. Remarkably, on some datasets, \gls{loe} trained on contaminated data can achieve better results than on clean data (as shown in Tables~\ref{tab:tabular_f1} and ~\ref{tab:tabular_auc}), suggesting that the latent anomalies provide a positive learning signal. Overall, we can see that \gls{loe} improves the performance of anomaly detection methods on contaminated tabular datasets significantly.
\begin{table*}[t!]
	\caption{F1-score ($\%$) with standard deviation for anomaly detection on 30 tabular datasets which are from the empirical study of \citet{shenkar2022anomaly}. For all experiments, we set the contamination ratio of the training set as $10\%$. The number in the brackets is the average performance difference from the model trained on clean data. 
	\gls{loe} outperforms the ``Blind'' and ``Refine'' baselines.}
	\label{tab:tabular_f1}
	\small
	\centering
	\vspace{3pt}
	\resizebox{\linewidth}{!}{
	\begin{tabular}{lcccc|cccc}
        \hline
        &\multicolumn{4}{c|}{NTL}&\multicolumn{4}{c}{ICL} \\
        &Blind& Refine& LOE$_H$ (ours)& LOE$_S$ (ours)&Blind& Refine& LOE$_H$ (ours)& LOE$_S$ (ours)\\
        \hline
\multirow{2}{*}{abalone}	& 37.9$\pm$13.4& 55.2$\pm$15.9& 42.8$\pm$26.9& \textbf{59.3$\pm$12.0}& 50.9$\pm$1.5& \textbf{54.3$\pm$2.9}& 53.4$\pm$5.2& 51.7$\pm$2.4\\ 
&(-25.3)&(-8.0)&(-20.4)& \textbf{(-3.9)}&(-11.2)& \textbf{(-7.8)}&(-8.7)&(-10.4)\\ 
\multirow{2}{*}{annthyroid}	& 29.7$\pm$3.5& 42.7$\pm$7.1& 47.7$\pm$11.4& \textbf{50.3$\pm$4.5}& 29.1$\pm$2.2& 38.5$\pm$2.1& \textbf{48.7$\pm$7.6}& 43.0$\pm$8.8\\ 
&(-21.6)&(-8.6)&(-3.6)& \textbf{(-1.0)}&(-12.0)&(-2.6)& \textbf{(+7.6)}&(+1.9)\\ 
\multirow{2}{*}{arrhythmia}	& 57.6$\pm$2.5& 59.1$\pm$2.1& 62.1$\pm$2.8& \textbf{62.7$\pm$3.3}& 53.9$\pm$0.7& 60.9$\pm$2.2& 62.4$\pm$1.8& \textbf{63.6$\pm$2.1}\\ 
&(-3.0)&(-1.5)&(+1.5)& \textbf{(+2.1)}&(-7.6)&(-0.6)&(+0.9)& \textbf{(+2.1)}\\ 
\multirow{2}{*}{breastw}	& 84.0$\pm$1.8& 93.1$\pm$0.9& \textbf{95.6$\pm$0.4}& 95.3$\pm$0.4& 92.6$\pm$1.1& 93.4$\pm$1.0& \textbf{96.0$\pm$0.6}& 95.7$\pm$0.6\\ 
&(-8.4)&(+0.7)& \textbf{(+3.2)}&(+2.9)&(-2.4)&(-1.6)& \textbf{(+1.0)}&(+0.7)\\ 
\multirow{2}{*}{cardio}	& 21.8$\pm$4.9& 45.2$\pm$7.9& \textbf{73.0$\pm$7.9}& 57.8$\pm$5.5& 50.2$\pm$4.5& 56.2$\pm$3.4& \textbf{71.1$\pm$3.2}& 62.2$\pm$2.7\\ 
&(-35.0)&(-11.6)& \textbf{(+16.2)}&(+1.0)&(-19.5)&(-13.5)& \textbf{(+1.4)}&(-7.5)\\ 
\multirow{2}{*}{ecoli}	& 0.0$\pm$0.0& 88.9$\pm$14.1& \textbf{100$\pm$0.0}& \textbf{100$\pm$0.0}& 17.8$\pm$15.1& 46.7$\pm$25.7& \textbf{75.6$\pm$4.4}& \textbf{75.6$\pm$4.4}\\ 
&(-95.6)&(-6.7)& \textbf{(+4.4)}&\textbf{(+4.4)}&(-55.5)&(-26.6)& \textbf{(+2.3)}&\textbf{(+2.3)}\\ 
\multirow{2}{*}{forest cover}	& 20.4$\pm$4.0& 56.2$\pm$4.9& 61.1$\pm$34.9& \textbf{67.6$\pm$30.6}& 9.2$\pm$4.5& 8.0$\pm$3.6& 6.8$\pm$3.6& \textbf{11.1$\pm$2.1}\\ 
&(-44.2)&(-8.4)&(-3.5)& \textbf{(+3.0)}&(-37.8)&(-39.0)&(-40.2)& \textbf{(-35.9)}\\ 
\multirow{2}{*}{glass}	& 11.1$\pm$7.0& 15.6$\pm$5.4& 17.8$\pm$5.4& \textbf{20.0$\pm$8.3}& 8.9$\pm$4.4& \textbf{11.1$\pm$0.0}& 11.1$\pm$7.0& 8.9$\pm$8.3\\ 
&(-6.7)&(-2.2)&(+0.0)& \textbf{(+2.2)}&(-13.3)& \textbf{(-11.1)}&(-11.1)&(-13.3)\\ 
\multirow{2}{*}{ionosphere}	& 89.0$\pm$1.5& 91.0$\pm$2.0& 91.0$\pm$1.7& \textbf{91.3$\pm$2.2}& 86.5$\pm$1.1& 85.9$\pm$2.3& 85.7$\pm$2.8& \textbf{88.6$\pm$0.6}\\ 
&(-3.5)&(-1.5)&(-1.5)& \textbf{(-1.2)}&(-5.7)&(-6.3)&(-6.5)& \textbf{(-3.6)}\\ 
\multirow{2}{*}{kdd}	& 95.9$\pm$0.0& 96.0$\pm$1.1& 98.1$\pm$0.4& \textbf{98.4$\pm$0.1}& 99.3$\pm$0.1& 99.4$\pm$0.1& \textbf{99.5$\pm$0.0}& 99.4$\pm$0.0\\ 
&(-2.4)&(-2.3)&(-0.2)& \textbf{(+0.1)}&(-0.1)&(+0.0)& \textbf{(+0.1)}&(+0.0)\\ 
\multirow{2}{*}{kddrev}	& 98.4$\pm$0.1& 98.4$\pm$0.2& 89.1$\pm$1.7& \textbf{98.6$\pm$0.0}& 97.9$\pm$0.5& 98.4$\pm$0.4& \textbf{98.8$\pm$0.1}& 98.2$\pm$0.4\\ 
&(+0.2)&(+0.2)&(-9.1)& \textbf{(+0.4)}&(-0.9)&(-0.4)& \textbf{(+0.0)}&(-0.6)\\ 
\multirow{2}{*}{letter}	& 36.4$\pm$3.6& 44.4$\pm$3.1& 25.4$\pm$10.0& \textbf{45.6$\pm$10.6}& 43.0$\pm$2.5& 51.2$\pm$3.7& \textbf{54.4$\pm$5.6}& 47.2$\pm$4.9\\ 
&(-11.0)&(-3.0)&(-22.0)& \textbf{(-1.8)}&(-15.5)&(-7.3)& \textbf{(-4.1)}&(-11.3)\\ 
\multirow{2}{*}{lympho}	& 53.3$\pm$12.5& 60.0$\pm$8.2& 60.0$\pm$13.3& \textbf{73.3$\pm$22.6}& 43.3$\pm$8.2& 60.0$\pm$8.2& 80.0$\pm$12.5& \textbf{83.3$\pm$10.5}\\ 
&(-20.0)&(-13.3)&(-13.3)& \textbf{(+0.0)}&(-40.0)&(-23.3)&(-3.3)& \textbf{(+0.0)}\\ 
\multirow{2}{*}{mammogra.}	& 5.5$\pm$2.8& 2.6$\pm$1.7& 3.3$\pm$1.6& \textbf{13.5$\pm$3.8}& 8.8$\pm$1.9& 11.4$\pm$1.9& 34.0$\pm$20.2& \textbf{42.8$\pm$17.6}\\ 
&(-21.3)&(-24.2)&(-23.5)& \textbf{(-13.3)}&(-14.0)&(-11.4)&(+11.2)& \textbf{(+20.0)}\\ 
\multirow{2}{*}{mnist tabular}	& 78.6$\pm$0.5& \textbf{80.3$\pm$1.1}& 71.8$\pm$1.8& 76.3$\pm$2.1& 72.1$\pm$1.0& 80.7$\pm$0.7& \textbf{86.0$\pm$0.4}& 79.2$\pm$0.9\\ 
&(-6.6)& \textbf{(-4.9)}&(-13.4)&(-8.9)&(-10.5)&(-1.9)& \textbf{(+3.4)}&(-3.4)\\ 
\multirow{2}{*}{mulcross}	& 45.5$\pm$9.6& \textbf{58.2$\pm$3.5}& \textbf{58.2$\pm$6.2}& 50.1$\pm$8.9& 70.4$\pm$13.4& 94.4$\pm$6.3& \textbf{100$\pm$0.0}& 99.9$\pm$0.1\\ 
&(-50.5)& \textbf{(-37.8)}&\textbf{(-37.8)}&(-45.9)&(-29.6)&(-5.6)& \textbf{(+0.0)}&(-0.1)\\ 
\multirow{2}{*}{musk}	& 21.0$\pm$3.3& 98.8$\pm$0.4& \textbf{100$\pm$0.0}& \textbf{100$\pm$0.0}& 6.2$\pm$3.0& \textbf{100$\pm$0.0}& \textbf{100$\pm$0.0}& \textbf{100$\pm$0.0}\\ 
&(-79.0)&(-1.2)& \textbf{(+0.0)}&\textbf{(+0.0)}&(-93.8)& \textbf{(+0.0)}&\textbf{(+0.0)}&\textbf{(+0.0)}\\ 
\multirow{2}{*}{optdigits}	& 0.2$\pm$0.3& 1.5$\pm$0.3& 41.7$\pm$45.9& \textbf{59.1$\pm$48.2}& 0.8$\pm$0.5& \textbf{1.3$\pm$1.1}& 1.2$\pm$1.0& 0.9$\pm$0.5\\ 
&(-24.7)&(-23.4)&(+16.8)& \textbf{(+34.2)}&(-62.4)& \textbf{(-61.9)}&(-62.0)&(-62.3)\\ 
\multirow{2}{*}{pendigits}	& 5.0$\pm$2.5& 32.6$\pm$10.0& 79.4$\pm$4.7& \textbf{81.9$\pm$4.3}& 10.3$\pm$4.6& 30.1$\pm$8.5& 80.3$\pm$6.1& \textbf{88.6$\pm$2.2}\\ 
&(-56.3)&(-28.7)&(+18.1)& \textbf{(+20.6)}&(-67.9)&(-48.1)&(+2.1)& \textbf{(+10.4)}\\ 
\multirow{2}{*}{pima}	& 60.3$\pm$2.6& 61.0$\pm$1.9& \textbf{61.3$\pm$2.4}& 61.0$\pm$0.9& 58.1$\pm$2.9& 59.3$\pm$1.4& \textbf{63.0$\pm$1.0}& 60.1$\pm$1.4\\ 
&(-1.2)&(-0.5)& \textbf{(-0.2)}&(-0.5)&(-2.2)&(-1.0)& \textbf{(+2.7)}&(-0.2)\\ 
\multirow{2}{*}{satellite}	& 73.6$\pm$0.4& 74.1$\pm$0.3& \textbf{74.8$\pm$0.4}& 74.7$\pm$0.1& 72.7$\pm$1.3& 72.7$\pm$0.6& \textbf{73.6$\pm$0.2}& 73.2$\pm$0.6\\ 
&(-1.0)&(-0.5)& \textbf{(+0.2)}&(+0.1)&(-2.1)&(-2.1)& \textbf{(-1.2)}&(-1.6)\\ 
\multirow{2}{*}{satimage}	& 26.8$\pm$1.5& 86.8$\pm$4.0& 90.7$\pm$1.1& \textbf{91.0$\pm$0.7}& 7.3$\pm$0.6& 85.1$\pm$1.4& 91.3$\pm$1.1& \textbf{91.5$\pm$0.9}\\ 
&(-65.2)&(-5.2)&(-1.3)& \textbf{(-1.0)}&(-82.0)&(-4.2)&(+2.0)& \textbf{(+2.2)}\\ 
\multirow{2}{*}{seismic}	& 11.9$\pm$1.8& 11.5$\pm$1.0& \textbf{18.1$\pm$0.7}& 17.1$\pm$0.6& 14.9$\pm$1.4& 17.3$\pm$2.1& 23.6$\pm$2.8& \textbf{24.2$\pm$1.4}\\ 
&(-0.6)&(-1.0)& \textbf{(+5.6)}&(+4.6)&(-3.0)&(-0.6)&(+5.7)& \textbf{(+6.3)}\\ 
\multirow{2}{*}{shuttle}	& 97.0$\pm$0.3& 97.0$\pm$0.2& \textbf{97.1$\pm$0.2}& 97.0$\pm$0.2& 96.6$\pm$0.2& 96.7$\pm$0.1& 96.9$\pm$0.1& \textbf{97.0$\pm$0.2}\\ 
&(+0.3)&(+0.3)& \textbf{(+0.4)}&(+0.3)&(-0.4)&(-0.3)&(-0.1)& \textbf{(+0.0)}\\ 
\multirow{2}{*}{speech}	& 6.9$\pm$1.2& 8.2$\pm$2.1& 43.3$\pm$5.6& \textbf{50.8$\pm$2.5}& 0.3$\pm$0.7& 1.6$\pm$1.0& \textbf{2.0$\pm$0.7}& 0.7$\pm$0.8\\ 
&(-2.6)&(-1.3)&(+33.8)& \textbf{(+41.3)}&(-4.1)&(-2.8)& \textbf{(-2.4)}&(-3.7)\\ 
\multirow{2}{*}{thyroid}	& 43.4$\pm$5.5& 55.1$\pm$4.2& \textbf{82.4$\pm$2.7}& \textbf{82.4$\pm$2.3}& 45.8$\pm$7.3& 71.6$\pm$2.4& \textbf{83.2$\pm$2.9}& 80.9$\pm$2.5\\ 
&(-34.4)&(-22.7)& \textbf{(+4.6)}&\textbf{(+4.6)}&(-31.4)&(-5.6)& \textbf{(+6.0)}&(+3.7)\\ 
\multirow{2}{*}{vertebral}	& 22.0$\pm$4.5& 21.3$\pm$4.5& 22.7$\pm$11.0& \textbf{25.3$\pm$4.0}& 8.9$\pm$3.1& 8.9$\pm$4.2& 7.8$\pm$4.2& \textbf{10.0$\pm$2.7}\\ 
&(-8.7)&(-9.4)&(-8.0)& \textbf{(-5.4)}&(-7.8)&(-7.8)&(-8.9)& \textbf{(-6.7)}\\ 
\multirow{2}{*}{vowels}	& 36.0$\pm$1.8& 50.4$\pm$8.8& \textbf{62.8$\pm$9.5}& 48.4$\pm$6.6& 42.1$\pm$9.0& 60.4$\pm$7.9& \textbf{81.6$\pm$2.9}& 74.4$\pm$8.0\\ 
&(-40.7)&(-26.3)& \textbf{(-13.9)}&(-28.3)&(-37.5)&(-19.2)& \textbf{(+2.0)}&(-5.2)\\ 
\multirow{2}{*}{wbc}	& 25.7$\pm$12.3& 45.7$\pm$15.5& \textbf{76.2$\pm$6.0}& 69.5$\pm$3.8& 50.5$\pm$5.7& 50.5$\pm$2.3& \textbf{61.0$\pm$4.7}& \textbf{61.0$\pm$1.9}\\ 
&(-39.1)&(-19.1)& \textbf{(+11.4)}&(+4.7)&(-8.2)&(-8.2)& \textbf{(+2.3)}&\textbf{(+2.3)}\\ 
\multirow{2}{*}{wine}	& 24.0$\pm$18.5& 66.0$\pm$12.0& 90.0$\pm$0.0& \textbf{92.0$\pm$4.0}& 4.0$\pm$4.9& 10.0$\pm$8.9& 98.0$\pm$4.0& \textbf{100$\pm$0.0}\\ 
&(-68.0)&(-26.0)&(-2.0)& \textbf{(+0.0)}&(-86.0)&(-80.0)&(+8.0)& \textbf{(+10.0)}\\ 
\hline
	\end{tabular}
	}
\end{table*}

\begin{table*}[t!]
	\caption{AUC ($\%$) with standard deviation for anomaly detection on 30 tabular datasets which are from the empirical study of \citet{shenkar2022anomaly}. For all experiments, we set the contamination ratio of the training set as $10\%$. The number in the brackets is the average performance difference from the model trained on clean data. 
	\gls{loe} outperforms the ``Blind'' and ``Refine'' baselines.}
	\label{tab:tabular_auc}
	\small
	\centering
	\vspace{3pt}
	\resizebox{\linewidth}{!}{
	\begin{tabular}{lcccc|cccc}
        \hline
        &\multicolumn{4}{c|}{NTL}&\multicolumn{4}{c}{ICL} \\
        &Blind& Refine& LOE$_H$ (ours)& LOE$_S$ (ours)&Blind& Refine& LOE$_H$ (ours)& LOE$_S$ (ours)\\
        \hline
\multirow{2}{*}{abalone}	& 91.4$\pm$1.7& 93.3$\pm$1.7& 93.4$\pm$1.0& \textbf{94.6$\pm$1.4}& 83.1$\pm$1.5& 91.2$\pm$0.8& 93.5$\pm$1.0& \textbf{93.6$\pm$0.8}\\ 
&(-2.4)&(-0.5)&(-0.4)& \textbf{(+0.8)}&(-10.1)&(-2.0)&(+0.3)& \textbf{(+0.4)}\\ 
\multirow{2}{*}{annthyroid}	& 66.1$\pm$2.8& 78.2$\pm$6.6& 83.9$\pm$7.0& \textbf{85.9$\pm$4.8}& 65.5$\pm$2.3& 73.1$\pm$2.5& \textbf{82.4$\pm$5.6}& 76.7$\pm$6.8\\ 
&(-19.1)&(-7.0)&(-1.3)& \textbf{(+0.7)}&(-8.7)&(-1.1)& \textbf{(+8.2)}&(+2.5)\\ 
\multirow{2}{*}{arrhythmia}	& 80.5$\pm$1.1& 82.5$\pm$0.8& 82.7$\pm$1.8& \textbf{84.8$\pm$1.7}& 75.5$\pm$0.3& 77.1$\pm$0.7& \textbf{79.2$\pm$0.2}& 78.4$\pm$0.8\\ 
&(-0.7)&(+1.3)&(+1.5)& \textbf{(+3.6)}&(-2.3)&(-0.7)& \textbf{(+1.4)}&(+0.6)\\ 
\multirow{2}{*}{breastw}	& 89.5$\pm$2.1& 96.1$\pm$0.8& \textbf{99.0$\pm$0.3}& 98.2$\pm$0.5& 97.1$\pm$0.8& 97.4$\pm$0.8& 98.7$\pm$0.3& \textbf{98.8$\pm$0.4}\\ 
&(-6.8)&(-0.2)& \textbf{(+2.7)}&(+1.9)&(-1.0)&(-0.7)&(+0.6)& \textbf{(+0.7)}\\ 
\multirow{2}{*}{cardio}	& 63.5$\pm$3.8& 76.9$\pm$3.8& \textbf{92.6$\pm$3.7}& 85.3$\pm$4.2& 80.0$\pm$1.4& 83.3$\pm$0.9& \textbf{91.1$\pm$1.9}& 87.5$\pm$2.1\\ 
&(-19.7)&(-6.3)& \textbf{(+9.4)}&(+2.1)&(-10.0)&(-6.7)& \textbf{(+1.1)}&(-2.5)\\ 
\multirow{2}{*}{ecoli}	& 74.9$\pm$8.2& 99.6$\pm$0.5& \textbf{100$\pm$0.0}& \textbf{100$\pm$0.0}& 80.4$\pm$4.2& 85.8$\pm$1.5& 88.5$\pm$1.8& \textbf{89.1$\pm$0.8}\\ 
&(-24.9)&(-0.2)& \textbf{(+0.2)}&\textbf{(+0.2)}&(-8.8)&(-3.4)&(-0.7)& \textbf{(-0.1)}\\ 
\multirow{2}{*}{forest cover}	& 91.2$\pm$2.2& \textbf{98.6$\pm$0.7}& 97.7$\pm$2.7& \textbf{98.6$\pm$2.1}& 73.0$\pm$11.7& 77.8$\pm$6.7& 78.9$\pm$3.2& \textbf{81.7$\pm$2.7}\\ 
&(-7.4)& \textbf{(+0.0)}&(-0.9)&\textbf{(+0.0)}&(-22.3)&(-17.5)&(-16.4)& \textbf{(-13.6)}\\ 
\multirow{2}{*}{glass}	& 75.1$\pm$4.0& 76.6$\pm$3.3& \textbf{77.8$\pm$4.8}& 77.1$\pm$4.6& 54.7$\pm$11.4& 66.6$\pm$5.7& 65.4$\pm$12.0& \textbf{71.5$\pm$9.2}\\ 
&(+2.6)&(+4.1)& \textbf{(+5.3)}&(+4.6)&(-25.9)&(-14.0)&(-15.2)& \textbf{(-9.1)}\\ 
\multirow{2}{*}{ionosphere}	& 95.6$\pm$0.8& \textbf{96.8$\pm$0.8}& 96.1$\pm$1.0& \textbf{96.8$\pm$0.9}& 92.6$\pm$1.1& 93.3$\pm$1.3& 88.7$\pm$3.3& \textbf{93.4$\pm$1.0}\\ 
&(-2.3)& \textbf{(-1.1)}&(-1.8)&\textbf{(-1.1)}&(-4.9)&(-4.2)&(-8.8)& \textbf{(-4.1)}\\ 
\multirow{2}{*}{kdd}	& \textbf{99.7$\pm$0.0}& 99.4$\pm$0.2& \textbf{99.7$\pm$0.0}& \textbf{99.7$\pm$0.0}& \textbf{99.9$\pm$0.0}& \textbf{99.9$\pm$0.0}& \textbf{99.9$\pm$0.0}& \textbf{99.9$\pm$0.0}\\ 
& \textbf{(-0.2)}&(-0.5)&\textbf{(-0.2)}&\textbf{(-0.2)}& \textbf{(+0.0)}&\textbf{(+0.0)}&\textbf{(+0.0)}&\textbf{(+0.0)}\\ 
\multirow{2}{*}{kddrev}	& \textbf{99.5$\pm$0.1}& 99.4$\pm$0.1& 96.1$\pm$0.9& \textbf{99.5$\pm$0.1}& 99.5$\pm$0.2& 99.7$\pm$0.1& \textbf{99.8$\pm$0.0}& 99.6$\pm$0.1\\ 
& \textbf{(+0.0)}&(-0.1)&(-3.4)&\textbf{(+0.0)}&(-0.3)&(-0.1)& \textbf{(+0.0)}&(-0.2)\\ 
\multirow{2}{*}{letter}	& 79.8$\pm$0.5& 83.5$\pm$0.8& 76.2$\pm$6.0& \textbf{84.3$\pm$4.8}& 82.3$\pm$2.9& 84.1$\pm$2.0& \textbf{86.2$\pm$2.8}& 83.7$\pm$2.0\\ 
&(-5.0)&(-1.3)&(-8.6)& \textbf{(-0.5)}&(-5.4)&(-3.6)& \textbf{(-1.5)}&(-4.0)\\ 
\multirow{2}{*}{lympho}	& 90.8$\pm$6.7& 93.7$\pm$3.2& 96.6$\pm$1.7& \textbf{98.1$\pm$2.2}& 94.1$\pm$2.0& 96.1$\pm$1.0& \textbf{98.9$\pm$1.0}& \textbf{98.9$\pm$1.1}\\ 
&(-6.3)&(-3.4)&(-0.5)& \textbf{(+1.0)}&(-5.3)&(-3.3)& \textbf{(-0.5)}&\textbf{(-0.5)}\\ 
\multirow{2}{*}{mammogra.}	& 68.7$\pm$6.2& 67.8$\pm$2.0& 69.2$\pm$3.8& \textbf{78.5$\pm$3.2}& 64.2$\pm$4.3& 69.7$\pm$4.7& 80.0$\pm$7.7& \textbf{84.0$\pm$4.3}\\ 
&(-13.8)&(-14.7)&(-13.3)& \textbf{(-4.0)}&(-14.8)&(-9.3)&(+1.0)& \textbf{(+5.0)}\\ 
\multirow{2}{*}{mnist tabular}	& 96.1$\pm$0.2& \textbf{96.7$\pm$0.4}& 94.7$\pm$0.5& 96.1$\pm$0.4& 94.1$\pm$0.4& 96.4$\pm$0.3& \textbf{97.9$\pm$0.1}& 96.3$\pm$0.2\\ 
&(-1.9)& \textbf{(-1.3)}&(-3.3)&(-1.9)&(-3.1)&(-0.8)& \textbf{(+0.7)}&(-0.9)\\ 
\multirow{2}{*}{mulcross}	& 81.7$\pm$7.5& \textbf{91.2$\pm$1.4}& 90.8$\pm$4.5& 82.6$\pm$10.5& 93.7$\pm$4.4& 99.4$\pm$0.7& \textbf{100$\pm$0.0}& \textbf{100$\pm$0.0}\\ 
&(-17.9)& \textbf{(-8.4)}&(-8.8)&(-17.0)&(-6.3)&(-0.6)& \textbf{(+0.0)}&\textbf{(+0.0)}\\ 
\multirow{2}{*}{musk}	& 76.2$\pm$2.3& \textbf{100$\pm$0.0}& \textbf{100$\pm$0.0}& \textbf{100$\pm$0.0}& 78.8$\pm$2.9& \textbf{100$\pm$0.0}& \textbf{100$\pm$0.0}& \textbf{100$\pm$0.0}\\ 
&(-23.8)& \textbf{(+0.0)}&\textbf{(+0.0)}&\textbf{(+0.0)}&(-21.2)& \textbf{(+0.0)}&\textbf{(+0.0)}&\textbf{(+0.0)}\\ 
\multirow{2}{*}{optdigits}	& 31.0$\pm$3.7& 38.7$\pm$3.8& 70.9$\pm$27.8& \textbf{72.6$\pm$33.6}& 13.8$\pm$4.2& \textbf{16.3$\pm$4.3}& 15.9$\pm$5.1& 14.6$\pm$3.7\\ 
&(-53.7)&(-46.0)&(-13.8)& \textbf{(-12.1)}&(-83.6)& \textbf{(-81.1)}&(-81.5)&(-82.8)\\ 
\multirow{2}{*}{pendigits}	& 64.0$\pm$9.3& 85.9$\pm$6.6& \textbf{99.1$\pm$0.5}& 98.9$\pm$0.4& 77.9$\pm$6.8& 83.3$\pm$4.7& 99.2$\pm$0.6& \textbf{99.7$\pm$0.1}\\ 
&(-33.1)&(-11.2)& \textbf{(+2.0)}&(+1.8)&(-21.3)&(-15.9)&(+0.0)& \textbf{(+0.5)}\\  
\multirow{2}{*}{pima}	& 59.5$\pm$3.4& 60.6$\pm$2.6& \textbf{60.8$\pm$1.8}& \textbf{60.8$\pm$1.0}& 58.2$\pm$3.7& 59.0$\pm$1.4& \textbf{64.1$\pm$1.5}& 61.1$\pm$1.4\\ 
&(-2.2)&(-1.1)& \textbf{(-0.9)}&\textbf{(-0.9)}&(-2.1)&(-1.3)& \textbf{(+3.8)}&(+0.8)\\ 
\multirow{2}{*}{satellite}	& 80.9$\pm$0.4& 82.2$\pm$0.3& 82.6$\pm$0.4& \textbf{82.9$\pm$0.3}& 78.5$\pm$1.2& 78.3$\pm$1.0& 79.3$\pm$0.9& \textbf{79.5$\pm$1.0}\\ 
&(-1.5)&(-0.2)&(+0.2)& \textbf{(+0.5)}&(-6.7)&(-6.9)&(-5.9)& \textbf{(-5.7)}\\ 
\multirow{2}{*}{satimage}	& 92.3$\pm$2.1& \textbf{99.7$\pm$0.1}& \textbf{99.7$\pm$0.1}& \textbf{99.7$\pm$0.1}& 89.8$\pm$1.6& 99.6$\pm$0.2& \textbf{99.7$\pm$0.1}& \textbf{99.7$\pm$0.1}\\ 
&(-7.5)& \textbf{(-0.1)}&\textbf{(-0.1)}&\textbf{(-0.1)}&(-9.9)&(-0.1)& \textbf{(+0.0)}&\textbf{(+0.0)}\\ 
\multirow{2}{*}{seismic}	& 51.6$\pm$0.5& 49.7$\pm$2.0& 50.3$\pm$3.0& \textbf{55.6$\pm$3.8}& 56.9$\pm$2.7& 58.4$\pm$2.3& \textbf{68.0$\pm$1.9}& 66.3$\pm$1.6\\ 
&(-1.3)&(-3.2)&(-2.6)& \textbf{(+2.7)}&(-6.5)&(-5.0)& \textbf{(+4.6)}&(+2.9)\\ 
\multirow{2}{*}{shuttle}	& 99.7$\pm$0.1& \textbf{99.8$\pm$0.1}& 99.7$\pm$0.1& 99.7$\pm$0.1& \textbf{99.7$\pm$0.1}& 99.6$\pm$0.0& \textbf{99.7$\pm$0.0}& \textbf{99.7$\pm$0.1}\\ 
&(+0.1)& \textbf{(+0.2)}&(+0.1)&(+0.1)& \textbf{(-0.3)}&(-0.4)&\textbf{(-0.3)}&\textbf{(-0.3)}\\ 
\multirow{2}{*}{speech}	& 48.6$\pm$1.2& 53.2$\pm$1.4& 78.8$\pm$3.0& \textbf{85.5$\pm$1.6}& 17.1$\pm$1.9& 21.8$\pm$1.5& \textbf{24.2$\pm$1.3}& 18.0$\pm$1.9\\ 
&(-13.9)&(-9.3)&(+16.3)& \textbf{(+23.0)}&(-41.3)&(-36.6)& \textbf{(-34.2)}&(-40.4)\\ 
\multirow{2}{*}{thyroid}	& 94.3$\pm$1.2& 96.4$\pm$0.3& 99.1$\pm$0.2& \textbf{99.3$\pm$0.2}& 96.0$\pm$0.9& 97.7$\pm$0.3& \textbf{99.4$\pm$0.2}& 99.2$\pm$0.3\\ 
&(-3.9)&(-1.8)&(+0.9)& \textbf{(+1.1)}&(-2.4)&(-0.7)& \textbf{(+1.0)}&(+0.8)\\ 
\multirow{2}{*}{vertebral}	& 54.8$\pm$4.6& 55.3$\pm$4.3& 47.9$\pm$12.0& \textbf{59.2$\pm$9.8}& 43.3$\pm$1.5& \textbf{50.5$\pm$2.7}& 45.6$\pm$5.7& 46.8$\pm$4.9\\ 
&(-5.0)&(-4.5)&(-11.9)& \textbf{(-0.6)}&(-10.5)& \textbf{(-3.3)}&(-8.2)&(-7.0)\\ 
\multirow{2}{*}{vowels}	& 87.6$\pm$2.2& 92.6$\pm$3.5& \textbf{96.3$\pm$1.9}& 92.7$\pm$2.7& 91.0$\pm$2.6& 95.6$\pm$2.0& \textbf{99.2$\pm$0.3}& 98.3$\pm$0.6\\ 
&(-10.4)&(-5.4)& \textbf{(-1.7)}&(-5.3)&(-7.9)&(-3.3)& \textbf{(+0.3)}&(-0.6)\\ 
\multirow{2}{*}{wbc}	& 81.2$\pm$7.0& 88.5$\pm$5.0& \textbf{94.9$\pm$2.2}& 93.4$\pm$2.4& 86.3$\pm$2.0& 86.8$\pm$1.1& \textbf{91.5$\pm$1.1}& 91.0$\pm$0.5\\ 
&(-11.6)&(-4.3)& \textbf{(+2.1)}&(+0.6)&(-4.6)&(-4.1)& \textbf{(+0.6)}&(+0.1)\\ 
\multirow{2}{*}{wine}	& 64.3$\pm$14.4& 93.1$\pm$7.7& 99.6$\pm$0.1& \textbf{99.8$\pm$0.1}& 49.9$\pm$12.6& 54.6$\pm$8.3& 99.7$\pm$0.7& \textbf{100$\pm$0.0}\\ 
&(-35.4)&(-6.6)&(-0.1)& \textbf{(+0.1)}&(-48.6)&(-43.9)&(+1.2)& \textbf{(+1.5)}\\ 
\hline
	\end{tabular}
	}
\end{table*}

\end{document}